\documentclass[10pt]{article}
\PassOptionsToPackage{final}{graphicx}
\usepackage{cosmosreport}
\usepackage{float}
\usepackage{placeins}
\usepackage{needspace}
\usepackage{graphicx}
\usepackage{longtable}
\usepackage{xspace}

\newcommand{\sysname}{\textsc{SWE-Prometheus}\xspace}

\newcommand{\NGI}{\textsc{NGI}\xspace}

\begin{document}

\reporttitleblock
{SWE-Prometheus: Measuring Engineering Governance\\[2pt]
Improvements in Real-World Repositories}
{{\small Jiajun Wu$^{1,*}$\hspace{0.7em}Leixin Sun$^{1,*}$\hspace{0.7em}Zihan Tan$^{1,*}$\hspace{0.7em}Yitao Liu$^{1}$\hspace{0.7em}Zhenjun Zhao$^{7}$\\[2pt]
 Danyang Zhang$^{8}$\hspace{0.7em}Shuo Li$^{5}$\hspace{0.7em}Jiaru Qian$^{2}$\hspace{0.7em}Yuxin Wu$^{6}$\hspace{0.7em}Shanghaoran Quan$^{2}$\\[2pt]
 Chuangxin Zhao$^{4}$\hspace{0.7em}Yangxu Liao$^{3}$\hspace{0.7em}Yang Liu$^{2}$\hspace{0.7em}Bin Chong$^{2,\ddagger}$\hspace{0.7em}Guancheng Wan$^{1,\ddagger}$\\[3pt]
 {\normalfont\small $^*$ Equal contribution \qquad $^\ddagger$ Corresponding author}}}
{$^1$ CosmosMind; $^2$ Peking University; $^3$ Tsinghua University; $^4$ HKUST; $^5$ ModCraft; $^6$ Renmin University of China; $^7$ University of Zaragoza; $^8$~ByteDance}
{\textbf{Large language model based coding agents have made substantial progress on repository-level software engineering tasks.} Existing repository benchmarks, however, usually start from a human-identified issue and evaluate whether a patch satisfies a functional signal. We present \textbf{\sysname}, a benchmark for the broader task of improving repository engineering governance. Each task provides a fixed snapshot and an open-ended objective, requiring the agent to identify risks, prioritize interventions, and verify the resulting changes. \sysname evaluates six governance dimensions through paired evidence, clean-environment probes, behavior gates, and two independent teacher ratings of the same evidence. The benchmark contains 60 repositories; ten models are evaluated on a shared 22-repository public subset, where mean Normalized Governance Improvement ranges from 0.0568 to 0.5760 and observed behavior-breakage rates range from 0\% to 23\%. On a frozen ten-repository batch, a repository-blind template obtains mean NGI 0.272, but its gains concentrate in Tests \& CI, Quality Gates, and Documentation; it improves Reproducible Environment and Dependency \& Security on none of the repositories. This baseline makes the distinction between adding governance artifacts and producing execution-backed improvements measurable. The no-op condition has median NGI zero and standard deviation 0.073; two teachers agree exactly on 57 of 60 dimension scores for the same no-op evidence. For the two highest conditional-mean systems, common-valid NGI is similar, while full-pool comparisons that include behavior failures favor Kimi-K3. These results show why repository-governance evaluation should report improvement, behavior preservation, evidence quality, and coverage together.
}
{\centering
 \linkbutton[accent]{\faHome}{Homepage}{https://cosmosmind.ai/leaderboard/swe-prometheus}\hspace{0.6em}
 \linkbutton[mOrange]{\faDatabase}{Hugging Face}{https://huggingface.co/datasets/CosmosMind/SWE-Prometheus}\hspace{0.6em}
 \linkbutton[posgreen]{\faGithub}{GitHub}{https://github.com/CosmosMind-ai/SWE-Prometheus}\par}

\section{Introduction}
\label{sec:intro}

\begin{figure}[t]
\centering
\includegraphics[width=\linewidth]{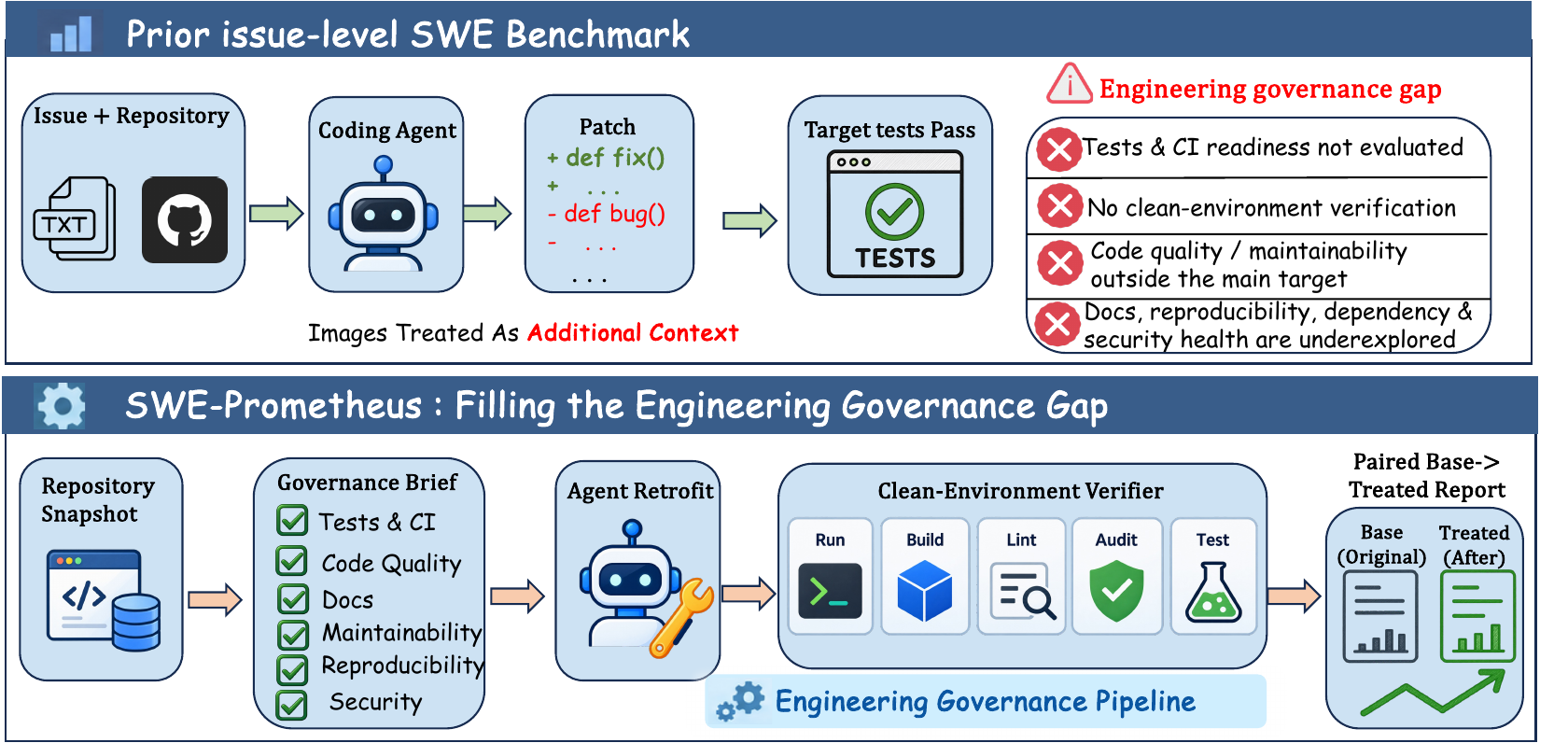}
\caption{Motivation for \sysname. Prior issue-level SWE benchmarks hand the agent a localized defect and score a patch against target tests, leaving test and CI readiness, clean-environment behavior verification, quality and maintainability outside the patch target, and documentation, reproducibility, dependency, and security health unevaluated. \sysname replaces the issue with a governance brief, requires the agent to diagnose and retrofit the repository itself, and scores a paired base-to-treated report produced by a clean-environment verifier.}
\label{fig:motivation}
\end{figure}

Large language model based coding agents have made substantial progress on repository-level software engineering. They can search unfamiliar codebases, execute development commands, edit multiple files, and produce end-to-end changes. Repository-level benchmarks such as SWE-bench made this progress measurable by pairing a real repository snapshot with a concrete issue and executable validation~\citep{jimenez2024swebench}. Agent scaffolds further showed that the interface through which a model searches, edits, and tests a repository can materially affect the outcome~\citep{yang2024sweagent}.

As illustrated in Figure~\ref{fig:motivation}, most existing evaluations begin after a human has already identified the problem. The issue statement specifies what the agent should investigate, while tests or a reference patch provide a relatively clear success signal. This is appropriate for targeted functional repair, but it leaves a central part of software engineering unmeasured: making a repository easier to test, maintain, reproduce, and safely evolve. In practice, these properties are distributed across tests, CI, tooling, documentation, project structure, environments, and dependencies. They rarely appear as one localized defect, and their improvement cannot be established from a diff alone.

The resulting gap is between engineering artifacts and engineering outcomes. An agent may add a workflow, a lint configuration, or a test file without making the repository more useful; it may also improve one dimension while introducing dependency drift or behavioral regressions elsewhere. A suitable evaluation must therefore give the agent room to diagnose the repository, compare evidence before and after intervention, and verify that improvements survive a clean execution environment.

To address this gap, we introduce \textbf{SWE-Prometheus}, a benchmark for autonomous repository-level engineering governance. Each task starts from a fixed repository snapshot and a general retrofit objective. The agent is not given a defect list, base score, hidden-test description, or target patch. It must inspect the repository, identify consequential gaps, prioritize interventions, and implement useful changes under a limited budget. The evaluator then reconstructs the base and treated states, executes the same probes in a clean environment, and applies behavior gates to identify unsafe interventions. SWE-Prometheus consequently evaluates whether an agent can leave a repository in a more trustworthy engineering state.

The benchmark covers six complementary governance dimensions: Tests \& CI, Code Quality Gates, Documentation \& Collaboration, Structure \& Maintainability, Reproducible Environment, and Dependency \& Security Health. Each dimension is scored from executable evidence and structured judgments, while applicability, unavailable evidence, gate strength, and behavior status remain explicit. The current release contains 60 real repositories, including 22 public instances shared across ten model rollouts and 38 private instances reserved for internal evaluation. On this shared public subset, mean Normalized Governance Improvement ranges from 0.0568 to 0.5760, while observed behavior-breakage rates range from 0\% to 23\%.

The auxiliary study shows how to interpret these differences. A repository-blind template reaches mean NGI 0.272, but its gains are concentrated in D1--D3; it improves D5 and D6 on none of the ten repositories. Patch-level evidence makes this boundary concrete: on Efficient-WAM, the template's workflow is configured to run Ruff and pytest, but its only test is \texttt{assert True}; the test passes while the separate Ruff probe still reports 117 errors. The baseline therefore calibrates how much score can come from standard governance artifacts and highlights why execution and failure exposure matter. The no-op condition has median NGI zero and standard deviation 0.073, while two teacher ratings agree exactly on 57 of 60 scores for the same no-op evidence. A three-pass maintainer review of the same sample of behavior-preserved patches found the changes useful and no new functional problems. Finally, the comparison between Kimi-K3 and GLM-5.3-Flash shows that improvement conditional on a successful patch and reliability across the full repository pool are distinct outcomes. Together, these controls make the benchmark's evidence requirements and interpretation explicit.

In summary, our contributions are threefold:

\begin{itemize}[leftmargin=1.8em,itemsep=1pt]
\item We identify repository-level engineering governance as a distinct evaluation setting that complements issue-level functional repair, and formulate it as an open-ended Repository Retrofit task without a defect list or oracle patch.
\item We introduce SWE-Prometheus, a repository-centered benchmark with six governance dimensions, paired base/treated evidence, clean-environment probes, behavior gates, public/private splits, and machine-readable release records.
\item We benchmark ten coding agents and establish diagnostic controls---including no-op, mechanical, rule-based, matched, and gate-strength analyses---that separate governance improvement from template effects, behavior risk, and verifier limitations.
\end{itemize}

\section{Related Work}
\label{sec:related}

\subsection{Code and Software Engineering Benchmarks}

The evaluation of code-generation capabilities has evolved from function-level correctness to increasingly realistic repository-level challenges. SWE-bench established the now-standard setting in which an agent receives a real issue, a repository snapshot, and executable tests~\citep{jimenez2024swebench}. Its focus on real GitHub repositories exposed the difficulty of understanding and modifying an unfamiliar codebase rather than completing an isolated function. SWE-bench Verified subsequently added human validation of task solvability and test quality~\citep{openai2024swebenchverified}, making the benchmark's success signal more reliable.

Earlier code-generation work established complementary function-level settings: HumanEval evaluates synthesized programs with unit tests, while MBPP covers short, crowd-sourced programming problems; surveys of machine learning for code situate these tasks within a broader history of code representation and generation~\citep{chen2021codex,austin2021mbpp,allamanis2018codesurvey}. These settings are useful for measuring local synthesis, but do not test repository-wide governance or maintenance. Other repository benchmarks broaden the repair target to visual issues and multilingual or multimodal issue resolution, including CodeV, OmniGIRL, and MM-IssueLoc~\citep{zhang2025codev,guo2025omnigirl,zhan2026mmissueloc}. Hardware repair and architectural code-smell repair further extend issue resolution to specialized targets~\citep{cui2026hwebench,dinu2026smellbench}; they remain issue-driven settings rather than open-ended assessments of repository readiness.

The repository-level setting has since expanded in several directions. SWE-bench-java, Multi-SWE-bench, and SWE-PolyBench extend issue resolution to Java, multiple programming languages, and broader repository distributions~\citep{zan2024swebenchjava,zan2025multiswebench,rashid2025swepolybench}. SWE-bench Multimodal and related visual benchmarks incorporate screenshots and other non-textual evidence~\citep{yang2024swebenchmultimodal}, while SWE-bench-Live and SWE-rebench study task freshness, automated collection, and contamination-aware evaluation~\citep{zhang2025swebenchlive,badertdinov2025swerebench}. SWE-bench Science and Rust-SWE-bench further suggest that domain-specific repositories and ecosystems expose failure modes hidden by a narrow benchmark distribution~\citep{xu2026swebenchscience,xiang2026rust}.

These benchmarks share several design principles with \sysname: fixed revisions, isolated execution, submitted patches, and executable checks. Their primary endpoint, however, remains the successful resolution of a known defect. \sysname changes the object of evaluation from repairing an identified issue to diagnosing and improving the engineering state of a repository.

\subsection{Software Engineering Agents}

The development of autonomous agents for repository-level software engineering has progressed alongside the benchmark ecosystem. SWE-agent shows that a purpose-built agent-computer interface can support effective search, editing, and testing~\citep{yang2024sweagent}. OpenHands provides a general platform for software-development agents~\citep{wang2024openhands}, while Agentless demonstrates that a deliberately simple localization, repair, and validation pipeline can remain competitive~\citep{xia2025agentless}. These studies highlight that agent performance depends not only on the underlying language model but also on how repository interaction is structured.

More general agent research studies reasoning interleaved with actions, learned tool use, and verbal feedback across interaction steps~\citep{yao2023react,schick2023toolformer,shinn2023reflexion}. WebArena evaluates agents in realistic browser-based environments, while OmniBench measures broad virtual-agent capabilities in a different task and environment scope~\citep{zhou2024webarena,bu2025omnibench}. Self-debugging work similarly examines how execution feedback can guide code revision~\citep{chen2023selfdebugging}. These methods inform the design of interactive coding agents, but are not themselves repository-governance benchmarks.

Other systems isolate complementary capabilities. RepoBench and RepoCoder focus on repository-level retrieval and code completion rather than issue resolution~\citep{liu2023repobench,zhang2023repocoder}. SWE-Explore treats repository exploration as a measurable capability~\citep{zhang2026sweexplore}, and RepairAgent and RepoAgent investigate autonomous program repair and repository-level development workflows~\citep{liu2024repairagent,wang2024repoagent}. Training-oriented work such as SWE-Gym uses real-world software engineering tasks to improve agent behavior~\citep{pan2025swegym}.

Visual-language research also studies multi-image perception and reasoning, with benchmarks and instruction-tuning resources such as MANTIS, MIRB, MuirBench, and OMIBench~\citep{jiang2024mantis,zhao2024mirb,wang2025muirbench,chen2026omibench}. These are adjacent to multimodal repository tasks because they examine visual evidence, but their general multi-image questions are not interchangeable with software issue localization or repository-level improvement.

Together, these systems motivate our separation of reconnaissance, intervention, and verification. They generally assume that the target issue, repair objective, or evaluation signal is already specified. SWE-Prometheus instead asks the agent to discover the engineering target itself, prioritize among repository weaknesses, and justify the intervention with evidence.

\subsection{Evaluation Beyond Patch Correctness}

Passing the original test suite is necessary for a useful patch, but it is not sufficient evidence of broader software engineering quality. RepoExec studies repository-level executable evaluation~\citep{lehai2025repoexec}, while SWT-Bench emphasizes validating generated fixes against real-world behavior~\citep{chen2024swtbench}. EvalPlus shows that strengthening test suites can substantially change conclusions about functional correctness~\citep{evalplus2023}. These results point to the importance of evaluation signals that are difficult to satisfy through narrow test overfitting.

Testing research provides additional tools for assessing evidence quality. Mutation testing estimates whether a test suite detects meaningful behavioral changes~\citep{jia2011mutation}. Empirical studies of flaky tests show that instability in the test infrastructure can itself contaminate engineering conclusions~\citep{luo2014flakytests}. Behavioral testing frameworks such as CheckList likewise argue for testing a range of observable behaviors rather than relying on a single aggregate score~\citep{ribeiro2020checklist}.

Classical diagnosis provides a further conceptual basis for separating observed failures from candidate explanations: Reiter formalizes diagnosis from first principles, and logical-abduction methods have been applied to diagnosing and correcting source-code design inconsistencies~\citep{reiter1987diagnosis,castro2011logicalabduction}. Our setting differs in that the agent must proactively identify governance gaps, rather than explain a supplied failure or inconsistency.

Our behavior gates and gate-strength labels build on this line of work by treating preservation evidence as a first-class outcome. A treated repository is not credited merely because a new check passes: the evaluation also asks whether existing behavior remains valid, whether the evidence is strong enough, and whether the run is reproducible.

\subsection{Software Quality and Engineering Governance}

The broader software quality literature offers useful concepts for describing repository health. ISO/IEC 25010 organizes quality around maintainability, reliability, security, and related properties~\citep{iso25010}. Complexity metrics~\citep{mccabe1976complexity,chidamber1994metrics}, technical-debt research~\citep{li2015technicaldebt,letouzey2012sqale}, and empirical studies of continuous integration~\citep{vasilescu2015quality,beller2017oops} provide practical measurement primitives for these dimensions.

Modern repository governance also depends on dependency scope, supply-chain provenance, and reproducible environments. OpenSSF Scorecard, SLSA, and NIST SSDF provide practical security and supply-chain guidance~\citep{openssfscorecard,slsa2023specification,nist2022ssdf}. Empirical work further shows that dependency alerts do not all have the same production relevance, making naive alert counts an inadequate proxy for risk~\citep{abdalkareem2022dependencies}.

These approaches typically audit a property, metric, or development practice in isolation. They do not evaluate whether an autonomous agent can identify the most consequential repository gaps, coordinate changes across several dimensions, and preserve existing behavior. In SWE-Prometheus, quality and governance tools are evidence probes inside a paired base-to-treated evaluation, not substitutes for the benchmark objective. The benchmark therefore evaluates whether an agent can turn engineering practices into durable repository improvements rather than merely adding configuration files.

\subsection{Reliable Benchmark Construction and Judge Calibration}

Benchmark design determines which capabilities can be measured reliably. Weak tests, leakage, incomplete curation, and unstable environments can all change model conclusions. Benchmark construction studies show that automated setup and broader repository coverage can produce task distributions that differ materially from popular curated suites~\citep{vergopoulos2025setupagent}. Contamination-limited evaluation and benchmark governance therefore require explicit provenance, release controls, and reproducible execution~\citep{kapoor2023leakage,li2024livebench,martinez2021benchmarks}.

Subjective dimensions introduce a second reliability challenge. LLM-as-a-judge methods can be useful, but they require calibration and agreement checks rather than unqualified trust~\citep{zheng2023judging,liu2023geval}. SWE-Prometheus follows these principles by retaining failed runs, measuring characterization-test strength through mutation testing, separating public and private data, and preserving per-run evidence. The remaining gap is an integrated evaluation in which the agent must discover the engineering target and improve it under realistic constraints while the evaluator checks both the intervention and the repository's continued usefulness.

\section{Dataset Overview}
\label{sec:method}

\begin{figure}[t]
\centering
\includegraphics[width=\linewidth]{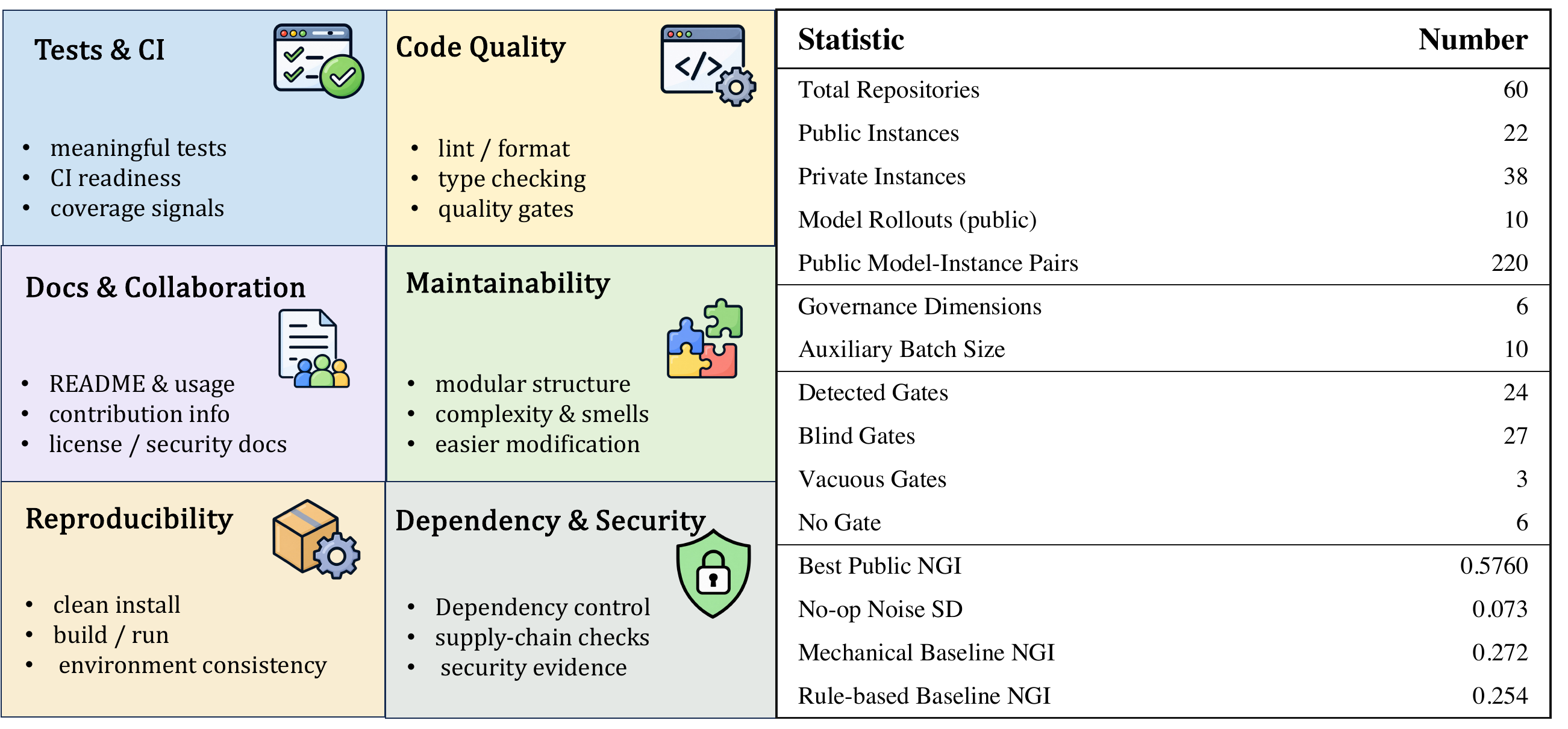}
\caption{Dataset and evaluation scope. The benchmark covers six governance dimensions and reports the public/private split, behavior-gate composition, and completed auxiliary-batch statistics.}
\label{fig:data-scope}
\end{figure}

\subsection{Task formulation and scope}
Each instance is a fixed snapshot of a real GitHub repository identified by \texttt{repo} and \texttt{base\_commit}. The agent receives a general instruction:

\begin{quote}
Inspect this repository and improve its engineering readiness across the applicable governance dimensions. Preserve existing behavior and public interfaces. Make useful changes for future users and maintainers, and verify important claims with executable checks.
\end{quote}

The task provides no issue list, base score, hidden test, or reference patch. Multiple governance strategies may be valid. The agent can inspect source code and public documentation, execute development tools, edit project files, and submit a unified patch, but cannot access evaluator evidence or other model results.

\paragraph{Governance dimensions.}
The six dimensions are scored from 1 to 5; a score of 4 denotes acceptable governance for the repository type and task scope. Figure~\ref{fig:data-scope} visualizes their scope: Tests \& CI, Code Quality Gates, Documentation \& Collaboration, Structure \& Maintainability, Reproducible Environment, and Dependency \& Security. Configuration presence alone is never sufficient evidence of improvement; each dimension must be supported by executable or structured evidence.

\paragraph{Evidence loop.}
The evaluator reconstructs the base repository, applies the agent patch to an independent treated snapshot, and runs the same installation, test, quality, complexity, dependency, and security probes on both states. Commands, exit codes, logs, environment versions, timeouts, and unavailable evidence are retained.

Each fixed base/treated evidence record is scored twice by independent teacher models using the same rubric and evidence. The released record retains both judgments, their aggregate, and their score spread. This provides a direct same-evidence inter-rater check; separately, the no-op condition measures end-to-end variation when an unchanged repository is processed through evidence collection and scoring.

\paragraph{Behavior gate.}
Characterization tests are validated on the base snapshot and then executed on the treated snapshot. A failed treated verification is labeled \texttt{behavior\_broken}; it is excluded from valid \NGI aggregation but retained in behavior-risk analysis. Mutation testing labels a gate as \texttt{detected}, \texttt{blind}, \texttt{vacuous}, or \texttt{none}.

\paragraph{Metrics.}
For instance $i$ and dimension $d$, raw improvement is $\Delta_{i,d}=s^{treated}_{i,d}-s^{base}_{i,d}$. Normalized Governance Improvement is
\[
NGI_i =
\frac{1}{|D_i|}\sum_{d\in D_i}
\frac{s^{treated}_{i,d}-s^{base}_{i,d}}
{5-s^{base}_{i,d}},
\]
where $D_i$ contains defined and scorable dimensions. A dimension with base score 5 receives no improvement reward but is checked for regression. We additionally report treated scores, valid and broken rates, no-regression rate, strict success, verified claim rate, and resource usage.

\section{Dataset Creation}
\label{sec:creation}

\subsection{Repository selection and screening}

SWE-Prometheus is intentionally repository-centered rather than issue-centered. We first collect candidate repositories with evidence of active development and at least one plausible governance gap. A candidate is retained only when (i) the base commit can be reconstructed, (ii) the project has meaningful executable behavior, (iii) a clean evaluation environment can be built without private credentials, (iv) at least two governance dimensions are applicable, and (v) the task can be described without prescribing a particular patch. Screening records language, build system, test entry points, dependency manager, license, commit age, and external-service requirements.

Discovery signals such as absent workflows, stale metadata, weak tests, and unpinned dependencies are used for high-recall candidate collection, not as final scores. The final score is based on executable evidence. Negative evidence is retained explicitly: an unavailable command, unsupported platform, or inapplicable dimension is recorded rather than silently treated as a zero-quality score.

\subsection{Evidence collection and task packaging}

For every retained repository, the construction process records a fixed repository identifier and base commit, a governance brief, applicable dimensions, baseline evidence, characterization commands, and environment metadata. The governance brief states the engineering objective without naming a defect or prescribing a patch. This preserves room for multiple valid retrofit strategies while keeping the task reproducible.

The evaluator reconstructs the base repository, runs installation and development probes, and stores commands, exit codes, logs, tool versions, timeouts, and unavailable evidence. The same evidence schema is used after treatment. This paired design prevents a configuration file from being counted as an improvement unless the corresponding command or structured check provides useful evidence.

\subsection{Execution protocol}

Every rollout starts from a fresh copy of the pinned base commit. The agent receives the same task instruction and repository snapshot for a given comparison set. The harness records tool calls, changed files, patch size, wall-clock time, token usage when exposed by the scaffold, and termination reason. After exit, the harness validates the patch and applies it to a second clean copy. Evaluation runs installation, characterization behavior, governance probes, and post-hoc artifact validation in that order.

Each command has a timeout, an exit-code policy, and an evidence schema. Logs are retained with tool versions and environment identifiers. Monetary cost is reported only when provider accounting is available; it is never imputed from incomplete accounting. This protocol separates infrastructure failures from engineering failures and makes each reported score auditable.

\subsection{Behavior gates and failure taxonomy}

Characterization tests are validated on the base snapshot and then executed on the treated snapshot. A failed treated verification is labeled \texttt{behavior\_broken}; it is excluded from valid \NGI aggregation but retained in behavior-risk analysis. Mutation testing labels a gate as \texttt{detected}, \texttt{blind}, \texttt{vacuous}, or \texttt{none}. The gate-strength label is reported alongside every behavior outcome because a zero-breakage result from a weak gate should not be interpreted as strong safety evidence.

We distinguish infrastructure failure, environment/dependency failure, invalid or empty patch, behavior regression, preserved behavior without governance improvement, and preserved behavior with positive improvement. A rollout may receive a primary failure label and secondary evidence flags. For example, a patch may preserve characterization behavior while introducing a vacuous CI gate; it should not be counted as a clean success merely because tests pass.

\subsection{Scoring and judge calibration}

Executable probes provide primary evidence for tests, CI, environment, and dependency dimensions. Documentation and maintainability use dimension-specific rubrics rather than a holistic impression. Each candidate score is accompanied by positive evidence, negative evidence, applicability, and a confidence flag. Two independent judgments are preferred; disagreements are adjudicated only after initial labels are preserved.

The calibration set is scored by independent evaluators who do not know model identity. We report agreement, adjudication rate, and sensitivity to rubric anchors, together with judge-family and prompt-order checks. These safeguards are part of the release protocol and are distinguished from completed benchmark results.

\section{Results}
\label{sec:exp}

\subsection{Experimental setting and dataset}

The current SWE-Prometheus release contains 60 real repository instances: 22 public instances shared by all current models and 38 private instances reserved for internal evaluation. Ten models have completed one rollout on the public subset, yielding 220 public model-instance pairs. The public subset is a shared comparison set rather than a random sample of all repositories.

Candidate repositories were discovered using signals such as missing or failing CI, thin tests, absent quality gates, incomplete collaboration documentation, unpinned dependencies, and maintenance gaps. These signals support high-recall discovery and stratification, while screening confirms meaningful functionality, a stable base commit, plausible governance headroom, and evaluation without private credentials or paid services.

The evaluation follows a paired base-to-treated design. For every model-instance pair, the harness first records baseline evidence, then applies the submitted patch to a clean copy and reruns the same probes. Characterization behavior is checked before governance scores are aggregated. We report valid and behavior-broken outcomes separately, retain unavailable evidence instead of converting it into a zero, and use the shared public subset for the primary cross-model comparison. Appendix~\ref{app:results} reports the detailed aggregate tables; the release package contains the machine-readable per-instance records used to reproduce them.

\subsection{Auxiliary conditions}

Model averages alone cannot separate capability from measurement artifact, so we additionally run a compact auxiliary study on ten repositories sampled from the public set with fixed strata for repository size, baseline quality, and gate availability. Every condition uses the same base snapshots and the same scoring pipeline, and all conditions reuse stored rollouts without issuing new model calls.

The batch includes a \emph{no-op} condition that pushes an empty patch through the full pipeline to measure end-to-end variation; a \emph{mechanical} retrofit that adds a fixed governance template without inspecting the repository; a \emph{rule-based} retrofit that first detects which artifacts are missing and adds only those, generating an import smoke test rather than a vacuous assertion; and the existing model rollouts evaluated on the same instances. A separate behavior-gate ablation uses all 60 instances because the public subset contains no \texttt{none} gate cases. Appendix~\ref{app:aux-design} documents the strata, the condition matrix, and the patch contents of each baseline.

\subsection{Governance improvement varies widely across agents}

\begin{figure}[!t]
\centering
\includegraphics[width=\linewidth]{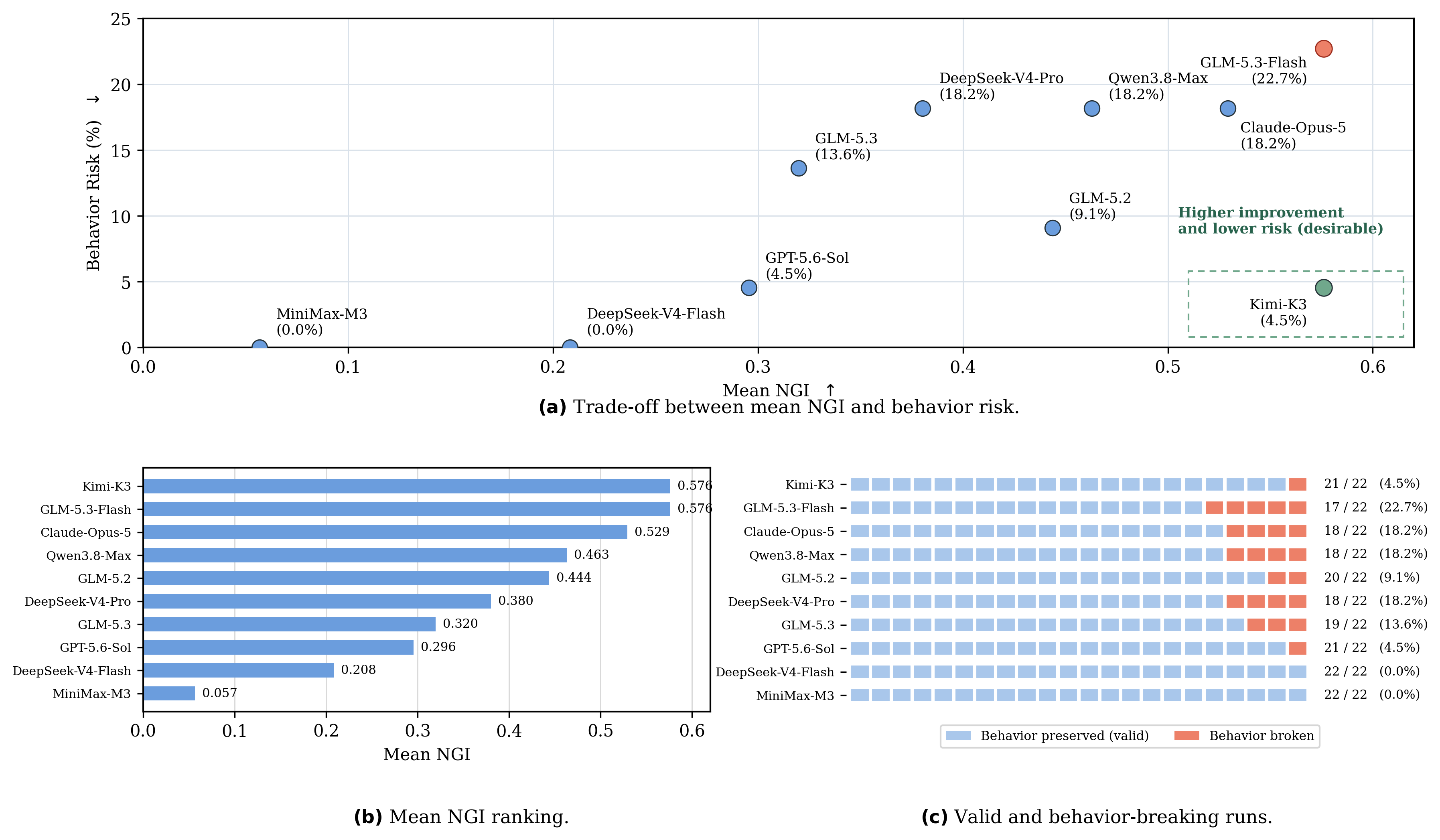}
\caption{Public model outcomes and behavioral validity. (a) Trade-off between mean NGI and behavior risk. (b) Conditional valid-run NGI comparison, not a reliability-aware ranking. (c) Per-model decomposition of valid and behavior-breaking runs across the same 22-instance pool. Claude-Opus-5 ran under Claude Code while the other nine models used pi.}
\label{fig:public-outcomes}
\end{figure}

Figure~\ref{fig:public-outcomes} reports the headline outcome on the public shared subset. Mean NGI spans roughly an order of magnitude, from 0.0568 to 0.5760, while measured behavior breakage spans 0\% to 22.7\%. Kimi-K3 and GLM-5.3-Flash obtain the same highest valid-run mean NGI, but they sit at opposite ends of the safety axis: Kimi-K3 breaks behavior on one of 22 instances while GLM-5.3-Flash breaks it on five. The equal conditional means therefore do not imply equal overall reliability. MiniMax-M3 produces little normalized improvement despite zero observed behavior breaks. The contrast is the reason improvement, safety, and absolute treated quality are reported as separate axes rather than folded into one leaderboard number. The companion coverage view separates valid runs from behavior-broken runs before any NGI aggregation, making clear that a high score supported by fewer valid runs should be interpreted cautiously.

\begin{table}[t]
\centering
\scriptsize
\setlength{\tabcolsep}{7pt}
\renewcommand{\arraystretch}{1.08}
\setlength{\aboverulesep}{0pt}
\setlength{\belowrulesep}{0pt}
\resizebox{0.96\textwidth}{!}{%
\begin{tabular}{llrrrr}
\toprule
\rowcolor{tabhead}
\textbf{Agent system} & \textbf{Scaffold} & \textbf{Valid} & \textbf{Broken} & \textbf{Breakage (\%)} & \textbf{NGI mean} \\
\midrule
\rowcolor{tabband} GPT-5.6-Sol & pi & 21 & 1 & 4.5 & 0.2956\\
\rowcolor{white} Claude-Opus-5 & Claude Code & 18 & 4 & 18.2 & 0.5293\\
\rowcolor{tabband} Kimi-K3 & pi & 21 & 1 & 4.5 & 0.5760\\
\rowcolor{white} GLM-5.3-Flash & pi & 17 & 5 & 22.7 & 0.5760\\
\rowcolor{tabband} Qwen3.8-Max & pi & 18 & 4 & 18.2 & 0.4630\\
\rowcolor{white} GLM-5.2 & pi & 20 & 2 & 9.1 & 0.4438\\
\rowcolor{tabband} DeepSeek-V4-Pro & pi & 18 & 4 & 18.2 & 0.3803\\
\rowcolor{white} GLM-5.3 & pi & 19 & 3 & 13.6 & 0.3198\\
\rowcolor{tabband} DeepSeek-V4-Flash & pi & 22 & 0 & 0.0 & 0.2083\\
\rowcolor{white} MiniMax-M3 & pi & 22 & 0 & 0.0 & 0.0568\\
\bottomrule
\end{tabular}
}
\caption{Public shared-subset results. Valid and broken counts are out of 22 instances. NGI is aggregated only over valid runs, which preserve characterization behavior and contain scorable evidence. Claude used Claude Code; the other models used pi.}
\label{tab:main}
\end{table}

\subsection{Reliability-aware comparison of the two leading systems}
\label{sec:reliability-comparison}

The valid-run mean is useful for measuring the quality of successful interventions, but it is conditional on behavior preservation. We therefore report two complementary comparisons for the two systems with the highest conditional mean. On the 17 repositories where both Kimi-K3 and GLM-5.3-Flash preserve characterization behavior, their mean NGI values are 0.547 and 0.576, respectively. The paired difference is $-0.029$ for Kimi-K3, with Kimi winning on 6 repositories, GLM-5.3-Flash on 10, and one tie. This difference is smaller than the empirical no-op standard deviation of 0.073 and does not support a reliable improvement advantage for either system.

To include delivery reliability, we also report a coverage-weighted NGI, defined as the mean over all 22 public repositories after assigning a behavior-broken run zero verified improvement. This is not a replacement for the conditional NGI; it asks a different practical question: how much verified improvement does a system deliver over the complete evaluation pool? Under this definition, Kimi-K3 obtains 0.550 and GLM-5.3-Flash obtains 0.445. We also examine a stronger penalty by assigning each behavior-broken run an NGI of $-1$; under this policy the means are 0.504 and 0.218, respectively. Finally, assigning the score-theoretic lower bound of $-3$ gives 0.413 for Kimi-K3 and $-0.237$ for GLM-5.3-Flash. The lower bound follows from the rubric's 1--5 scale and NGI normalization for an applicable dimension (base score 4, treated score 1). Across these full-pool policies Kimi-K3 remains first, although the ordering of the other systems changes substantially (Appendix~\ref{app:results}). Conditional means alone do not establish a reliable improvement difference between Kimi-K3 and GLM-5.3-Flash; their clearer distinction is delivery reliability, with one versus five behavior-broken repositories.

\begin{table}[t]
\centering
\small
\setlength{\tabcolsep}{5pt}
\renewcommand{\arraystretch}{1.08}
\begin{tabular}{@{}l r r r r l@{}}
\toprule
\headrow \textbf{Comparison} & \textbf{Repos.} & \textbf{Kimi} & \textbf{GLM-F} & \textbf{Difference} & \textbf{Interpretation}\\
\midrule
Common valid runs & 17 & 0.547 & 0.576 & $-0.029$ & Below no-op SD\\
\bandrow Full pool, broken $=0$ & 22 & 0.550 & 0.445 & $+0.105$ & Kimi more reliable\\
\bottomrule
\end{tabular}
\caption{Reliability-aware comparison of Kimi-K3 and GLM-5.3-Flash. ``GLM-F'' denotes GLM-5.3-Flash. The first row compares only repositories valid for both systems; the second retains all 22 repositories and assigns zero verified improvement to a behavior-broken treatment.}
\label{tab:reliability-comparison}
\end{table}

\subsection{Instance-level dispersion is large relative to the model spread}

Aggregates hide how unevenly governance improvement is distributed. Figure~\ref{fig:heterogeneity} shows the per-instance NGI distribution behind each mean. Every model spans a wide interquartile range, several have valid instances at or below zero, and mean and median diverge substantially for models whose gains are concentrated in a few repositories---Claude-Opus-5 in particular has a median of 0.722 against a mean of 0.457 on the auxiliary batch, indicating a right-skewed profile driven by a subset of instances. Because each model-instance pair has been run once, this dispersion is a joint effect of repository heterogeneity, rollout variability, and scorer noise, and it is what limits the resolution of the current comparison (Section~\ref{sec:sensitivity}).

\subsection{Diagnostic result analysis}
\label{sec:analysis}

This section reports the auxiliary conditions. Each subsection asks whether a specific part of the headline result survives a control.

\subsection{Non-agent controls and no-op measurement variation}
\label{sec:controls}

The no-op condition has mean NGI $-0.009$ and median 0.000, with five of ten instances receiving a non-zero score change and a standard deviation of 0.073. To measure judge disagreement separately, we compare the two independent teacher ratings on the same no-op evidence: they agree exactly on 57 of 60 repository-dimension scores, with non-zero disagreement in only three cells and a mean absolute score spread of 0.067 across the 60 cells. The median no-op NGI of zero indicates no systematic reward for an unchanged repository; the 0.073 standard deviation captures end-to-end variation, including evidence collection and scoring, rather than isolating scorer-only variance. These results show high agreement on fixed evidence alongside measurable pipeline variation. Differences of this size should be interpreted cautiously. Drift occurs in Structure \& Maintainability (three instances), Documentation, and Reproducible Environment (two each). The latter dimension is sensitive to installation evidence, so runtime variability is one plausible contributor; the structural drift on an unchanged diff is more directly consistent with judge variability. The no-op results do not identify the exact source of every changed score.

The two template conditions sit inside the model range rather than below it. The mechanical baseline obtains mean NGI 0.272 and the rule-based baseline 0.254, both above DeepSeek-V4-Flash and MiniMax-M3. The per-dimension decomposition clarifies what these scores represent: both baselines improve D1, D2, and D3 on nearly every repository, but improve D5 and D6 on none and D4 on only 20\% of repositories. In this batch, they capture the low-cost benefit of adding governance artifacts, while the measured evidence in D5 and D6 does not improve under these repository-blind or rule-based interventions. The ordering between the two baselines is itself informative: the deterministic rule baseline, which conditions its patch on detected repository state, does \emph{not} outperform a fixed template that ignores the repository entirely. Their separation, 0.018, is small relative to the observed no-op variation.

The patch-level evidence shows both why the template earns credit and where that credit stops. On Efficient-WAM, the mechanical template adds a GitHub Actions workflow that runs Ruff and pytest, plus a smoke test whose only assertion is \texttt{assert True}. The evaluator collects and passes that one test, but the same treated snapshot still reports 117 Ruff errors, no recognized install/build entry point, and no dependency manifest. This is a concrete example of artifact presence improving the test/CI-facing score without demonstrating meaningful behavior coverage or reproducible installation; it is consistent with the template's zero improvement on D5 and D6. We therefore interpret its NGI of 0.272 as a useful stress test of the rubric's artifact-facing dimensions, not as evidence that a repository-blind patch delivers the same engineering value as a targeted intervention.

\begin{figure}[!t]
\centering
\includegraphics[width=0.95\linewidth]{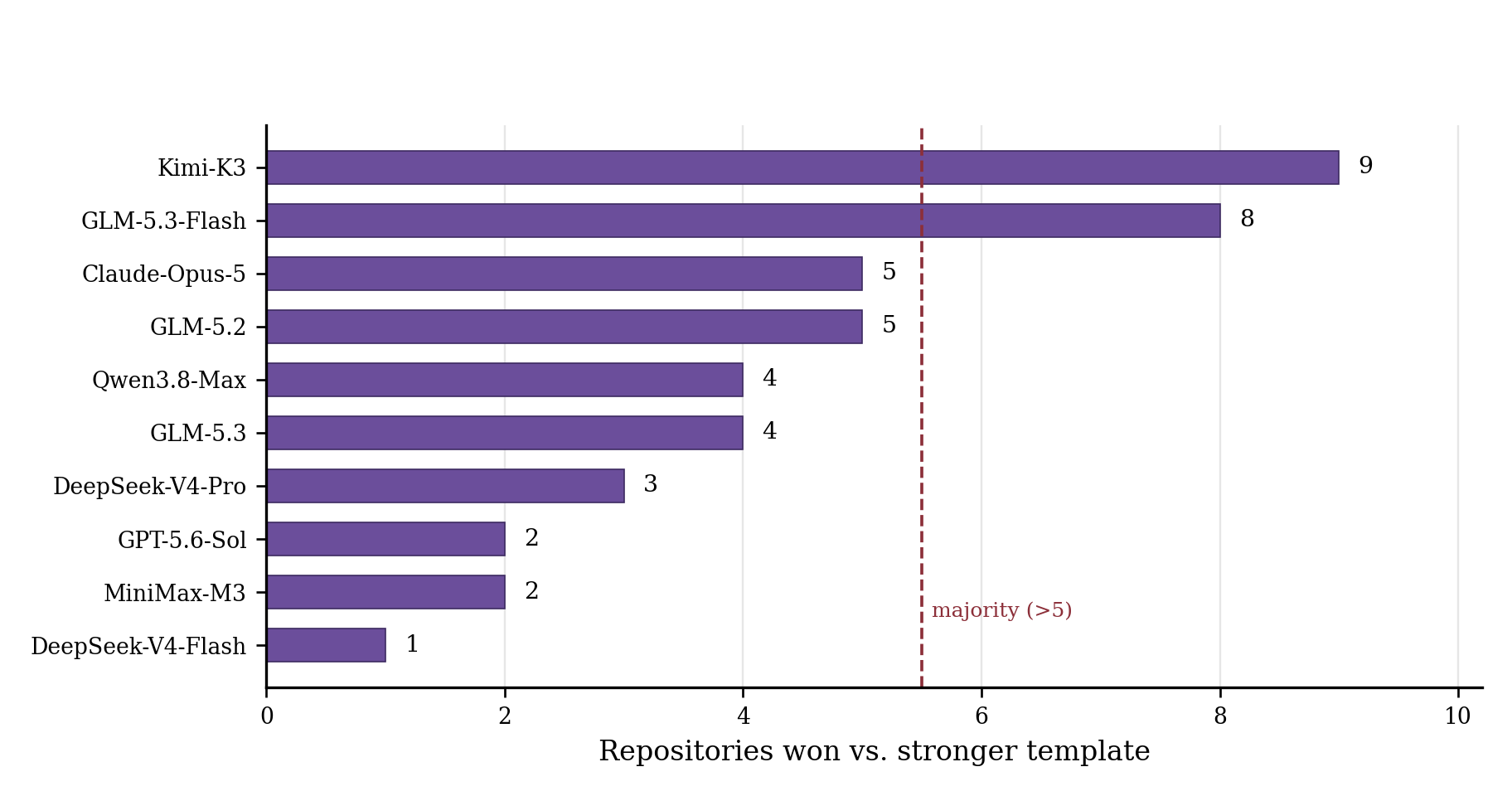}
\caption{Matched per-repository wins against the stronger template baseline. Each bar shows the number of repositories on which a model outperforms the stronger of the two template baselines on the frozen 10-repository batch. A majority requires more than 5 wins. Only Kimi-K3 and GLM-5.3-Flash clearly outperform the template on a majority of repositories.}
\label{fig:matched}
\end{figure}

\subsection{Matched comparison against the template baselines}
\label{sec:matched}

Comparing means across conditions conflates repository difficulty with agent capability, because an agent system and a baseline may succeed on different repositories. We therefore compare each agent system to the stronger of the two templates \emph{per instance}, counting an instance as a win only when the gap exceeds the empirical no-op noise floor. Figure~\ref{fig:matched} reports the resulting outcome switches.

The matched view is considerably less flattering than the mean ranking. Only Kimi-K3 (9 wins, 1 loss) and GLM-5.3-Flash (8--1) beat the template on a clear majority of repositories. Claude-Opus-5 and Qwen3.8-Max are close to even at 5--4 and 4--4. GPT-5.6-Sol is indistinguishable from the template on half of the batch, and the two weakest models are behind it: DeepSeek-V4-Flash loses on six of ten repositories and MiniMax-M3 on seven. In other words, most of the middle of the agent-system comparison is not reliably doing better, repository by repository, than a patch that never reads the repository.

\subsection{Verifier quality control}
\label{sec:verifier}

\begin{figure}[!t]
\centering
\begin{subfigure}{0.98\linewidth}
\centering
\includegraphics[width=\linewidth]{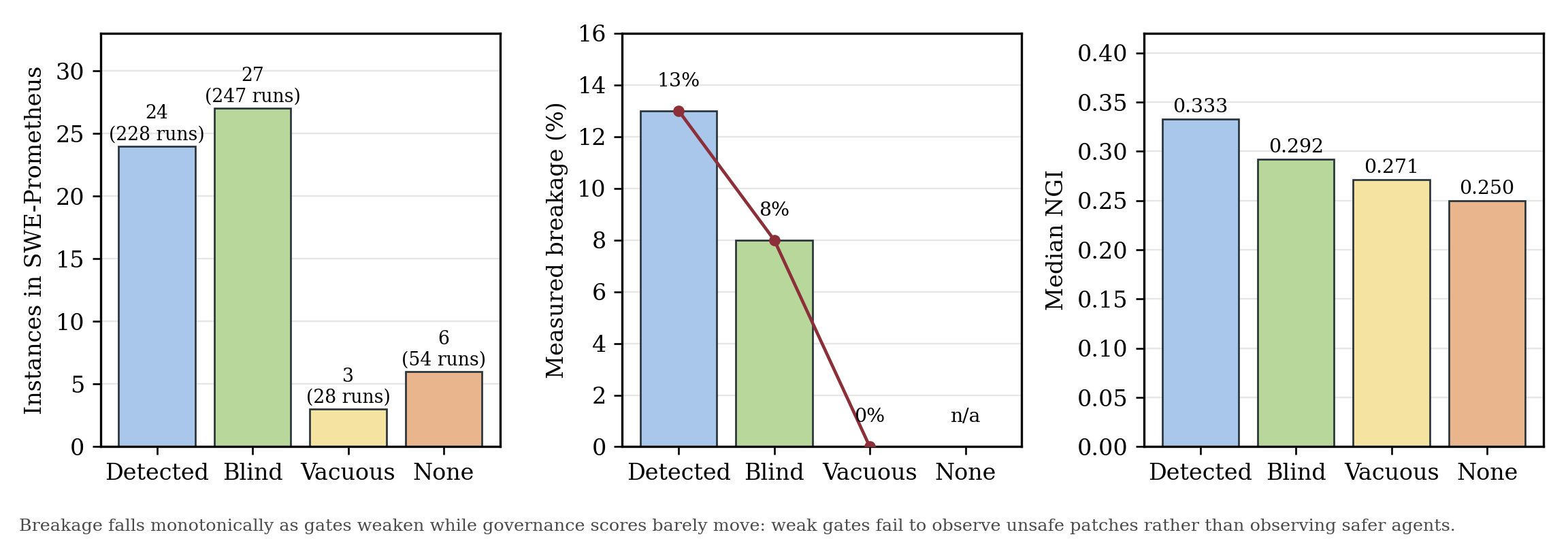}
\caption{Verifier quality control by mutation-testing gate strength.}
\label{fig:verifier}
\end{subfigure}
\\[-2pt]
\begin{subfigure}{0.98\linewidth}
\centering
\includegraphics[width=\linewidth]{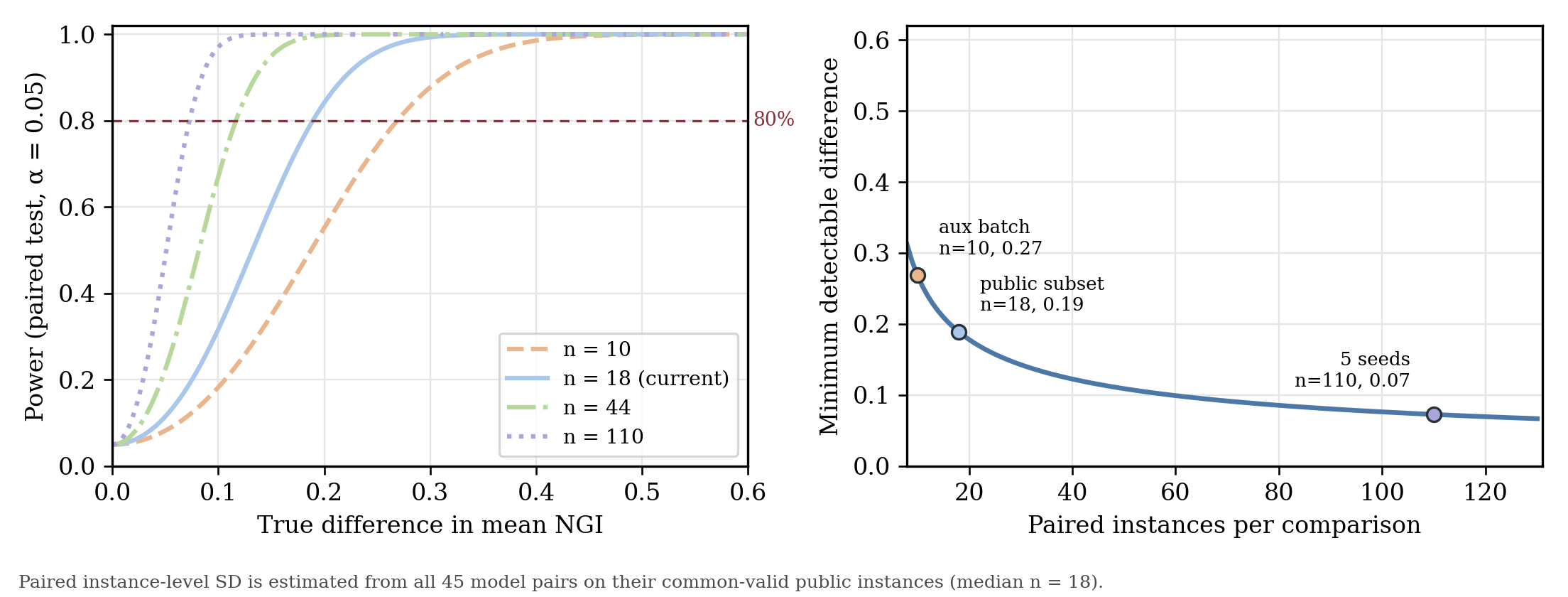}
\caption{Resolution of the current evaluation design.}
\label{fig:sensitivity}
\end{subfigure}
\caption{Verifier quality and statistical resolution. Weak gates observe fewer unsafe patches, while the current public subset has a finite resolution for detecting NGI differences.}
\label{fig:measurement-controls}
\end{figure}

A behavior gate is only as informative as it is discriminative, so we stratify outcomes by the mutation-testing label. Figure~\ref{fig:verifier} shows the result over all 60 instances and ten models. In this mutation-tested sample, measured breakage falls as gate strength degrades---13\% under detected gates, 8\% under blind gates, and 0\% under vacuous gates---while median NGI moves by only 0.06 across the same strata. This pattern is consistent with weaker gates missing regressions rather than proving that agents are safer. In particular, zero observed breakage under a gate known to miss the injected mutations is not evidence of preserved behavior. Headline breakage rates should therefore be reported with gate-strength breakdowns.

As a complementary validity check, we reviewed the same fixed sample of representative patches that had passed their behavior checks in three review passes, each time examining the patch together with its base and treated evidence. Across all three passes, the changes were judged useful for repository maintenance, and no new functional problems were identified. Repeating the review on the same sample checks whether the qualitative finding is stable rather than dependent on a single inspection.

The gate ablation itself is more nuanced than a simple inflation correction. Removing the gate and scoring every instance changes the median by at most 0.021 in either direction: it is positive for Kimi-K3 ($+0.014$), Qwen3.8-Max ($+0.021$), and GPT-5.6-Sol ($+0.021$), negative for GLM-5.3-Flash ($-0.017$), GLM-5.2 ($-0.008$), and GLM-5.3 ($-0.007$), and unchanged for the rest; the model ordering is identical with the gate on and off. Instances caught by the gate therefore tend to score near or below their model's average rather than above it, which suggests that behavior breakage is more often a symptom of a botched intervention than the price of an aggressive high-scoring one. The gate remains necessary---it identifies which scores are untrustworthy---but its effect should be reported as a sensitivity analysis, not assumed to be a monotone bias.

\subsection{Sensitivity of the current design}
\label{sec:sensitivity}

Given the dispersion in Figure~\ref{fig:heterogeneity}, we ask what difference the current design can actually resolve. Across all 45 model pairs on their common-valid public instances, the paired instance-level NGI difference has median standard deviation 0.270 over a median of 18 shared instances. Figure~\ref{fig:sensitivity} converts this into detection power. At 80\% power and $\alpha=0.05$, the public subset resolves a true mean difference of about 0.19, and the ten-repository auxiliary batch about 0.27.

This has a direct consequence for how Figure~\ref{fig:public-outcomes} should be read. Adjacent models in the ranking---Kimi-K3 and GLM-5.3-Flash, or GLM-5.2 and DeepSeek-V4-Pro---are separated by less than the resolution limit and are not distinguished by this instrument. The result supports a coarse grouping into a high band, a middle band, and a low band rather than a ten-way ordering. Reaching a resolution of 0.10 requires about 110 paired observations, or roughly five repeated rollouts per model-instance pair on the current public subset; this is the concrete cost of the repeated-seed experiment we defer to future work.

\subsection{Per-dimension decomposition}
\label{sec:perdim}

\begin{figure}[!t]
\centering
\begin{subfigure}{0.49\linewidth}
\centering
\includegraphics[width=\linewidth,height=0.742\linewidth]{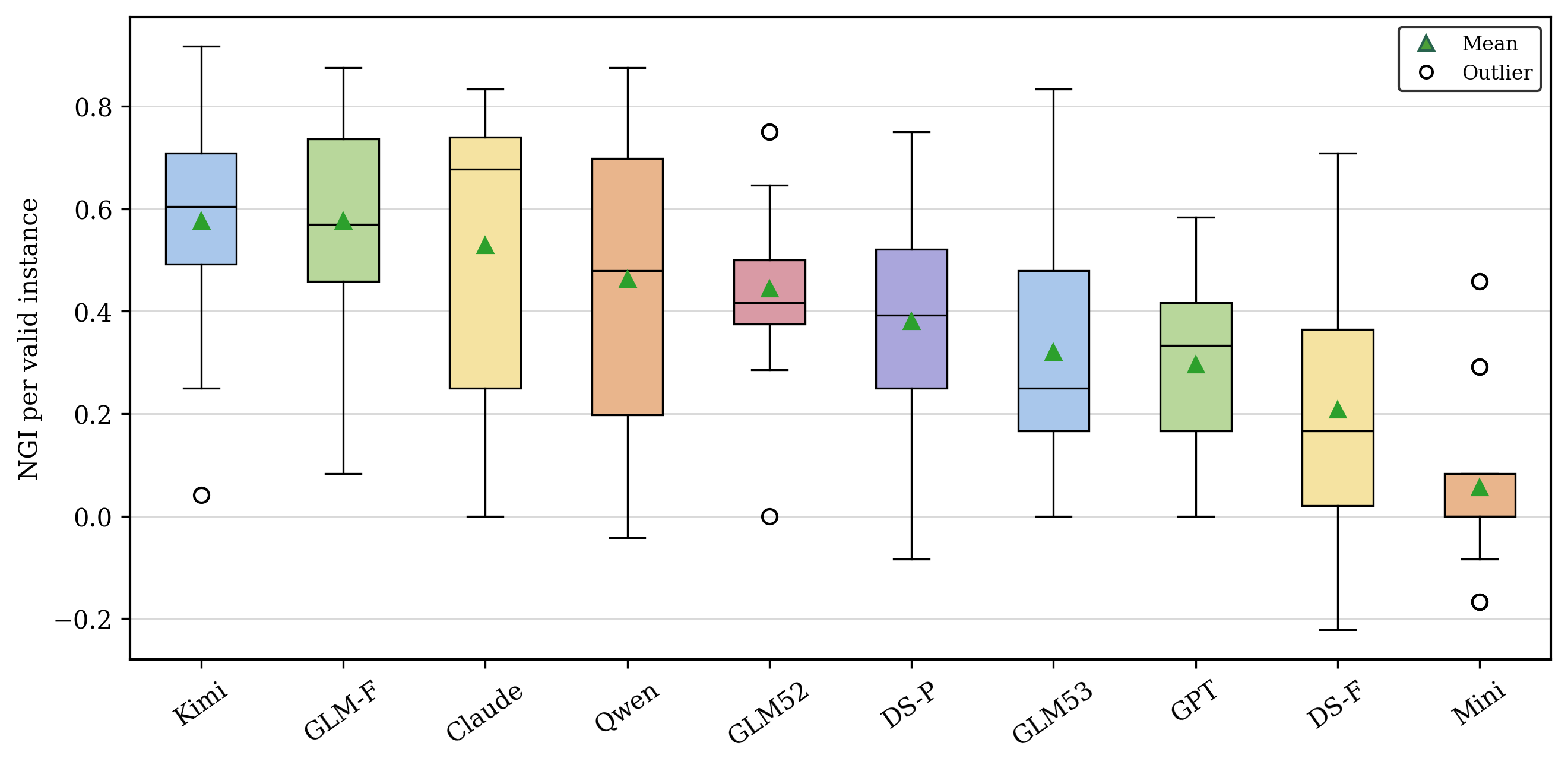}
\caption{Instance-level NGI distributions.}
\label{fig:ngi-distribution}
\end{subfigure}
\hfill
\begin{subfigure}{0.49\linewidth}
\centering
\includegraphics[width=\linewidth]{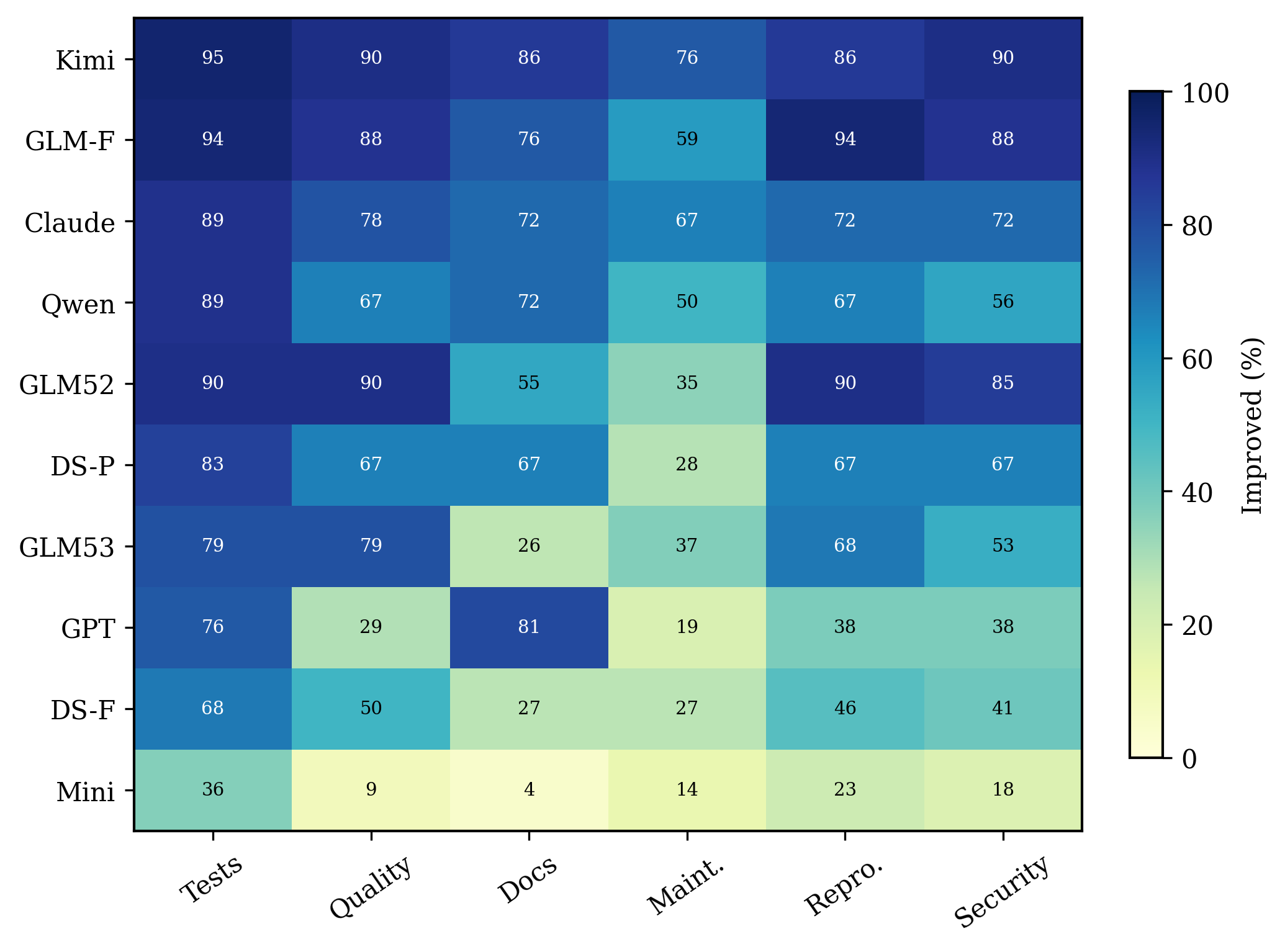}
\caption{Per-dimension improvement coverage.}
\label{fig:dimension-heatmap}
\end{subfigure}
\caption{Model heterogeneity and dimension coverage. The paired views show both the spread of repository-level gains and the governance dimensions improved by each model.}
\label{fig:heterogeneity}
\end{figure}

Aggregate NGI can hide a condition that improves one dimension while neglecting another. Figure~\ref{fig:heterogeneity} decomposes improvement across the six governance dimensions. Two patterns are consistent across models. Tests \& CI improves on the large majority of valid runs for every model except MiniMax-M3, whereas Structure \& Maintainability is the weakest or joint-weakest dimension for eight of ten models, falling to 19\% for GPT-5.6-Sol and 14\% for MiniMax-M3. Only Kimi-K3 (76\%) and Claude-Opus-5 (67\%) reach it on a clear majority of their valid runs, and Claude-Opus-5 is notable for reaching it while scoring lower than several competitors on the configuration-facing dimensions---the profile of a model that edits structure rather than adding files.

The same decomposition applied to the template baselines helps explain their standing in Section~\ref{sec:matched}. On this batch, both templates improve Tests \& CI, Code Quality Gates, and Documentation \& Collaboration on nearly every instance, and improve Reproducible Environment and Dependency \& Security Health on none, with Structure \& Maintainability at 20\%. This pattern is consistent with a distinction between dimensions that can receive credit from configuration or content and dimensions whose evidence depends on repository execution or dependency analysis. It does not imply that configuration-facing dimensions are inherently superficial; rather, the Efficient-WAM example shows why file presence alone is weaker evidence than successful, meaningful checks.

\subsection{Key insight: governance improvement is not repository diagnosis}
\label{sec:key-insight}

Taken together, the controls show that some measured governance gains can come from adding visible artifacts without repository-specific diagnosis. A fixed template that never reads the project reaches mean NGI 0.272 and exceeds several models on the frozen ten-repository batch; a rule-based patch that does inspect repository state scores 0.018 lower than that template, a difference small relative to the observed no-op variation. Meanwhile, the model/template gap is not uniform across dimensions: repository-specific execution evidence provides an important distinction, but these results do not establish that it is the only capability separating stronger systems.

The result is a measurement warning rather than evidence that all artifact-facing scores are uninformative: the Efficient-WAM example shows that file presence and functional evidence can diverge, while the dimension-level results identify where this template did and did not earn credit. Future versions should strengthen execution-backed evidence requirements for the template-reachable dimensions and report template baselines alongside model scores.

\section{Analysis}
\label{sec:discussion}

\subsection{What the benchmark makes visible}

The benchmark occupies a middle ground between issue resolution and static repository inspection. An issue benchmark asks whether a particular defect is fixed; a repository-health audit asks whether a project exhibits a property at one point in time. \sysname asks an agent to diagnose a project-wide gap, intervene under an open-ended objective, and provide evidence that the intervention is useful and behavior-preserving. The auxiliary controls make the two parts of this task visible: governance artifacts can be added mechanically, while repository-specific engineering outcomes require evidence from execution and inspection.

The central distinction the evidence loop enforces is between engineering \emph{presence} and engineering \emph{effectiveness}. A repository can contain a test suite, CI workflow, lint configuration, or dependency file without obtaining the intended benefit. The loop tests whether the associated command runs, whether its scope is meaningful, and whether the treated repository remains behaviorally compatible. Section~\ref{sec:perdim} shows that the mechanical template raises D1--D3 but does not improve D5 or D6 and rarely changes D4. The template is therefore a calibration point: it identifies where artifact-oriented scoring is most accessible and where executable evidence demands more than adding conventional files.

The six-dimensional view also exposes asymmetric progress. Tests and environment files are often easy to add, whereas dependency security, maintainability, and collaboration practices require understanding project intent. Reporting per-dimension intervention rates rather than aggregate NGI alone is therefore not optional presentation detail; it is what prevents a template-shaped result from being read as a diagnosis-shaped one.

\subsection{Improvement, safety, and absolute quality are separate axes}

Higher NGI does not automatically imply lower risk. An agent that makes many changes may improve more dimensions and also create more opportunities for regressions. Conversely, a zero-breakage model may simply make few changes---MiniMax-M3 has no observed breaks and the lowest improvement in the release. The Kimi-K3/GLM-5.3-Flash comparison makes the distinction concrete: their valid-run means are indistinguishable at the current resolution, but Kimi-K3 has the stronger full-pool verified outcome because it breaks fewer repositories. The combination of NGI, behavior-breakage rate, coverage-weighted improvement, no-regression rate, and verified claim rate is therefore more informative than any single leaderboard number.

The gate analysis in Section~\ref{sec:verifier} refines this. Removing the behavior gate raises the median for some models and lowers it for others rather than producing a universal inflation term, and the model ordering is unchanged. The gate's value is not that it corrects scores but that it marks which scores cannot be trusted, and gate strength determines how much confidence a zero-breakage observation can carry at all.

\subsection{Implications for agent design}

The task suggests three capabilities that are not captured by patch pass rates alone: repository reconnaissance, risk-aware intervention, and evidence-oriented verification. The matched comparison in Section~\ref{sec:matched} shows that only two of ten systems outperform the stronger template on a clear majority of the ten matched repositories. This makes repository-specific diagnosis a discriminating capability in the benchmark: aggregate gains alone do not show that an agent selected a better intervention than a repository-independent template.

Agents should therefore construct a compact repository map first, prioritize the gaps whose repair requires the project to actually build and run, and avoid changing stable source behavior when configuration or documentation is sufficient. They should then verify both positive claims and negative side effects. This motivates future baselines with explicit diagnosis, planning, patching, and audit phases, and it suggests that diagnosis quality---not patch volume---is the capability the next iteration of this benchmark should isolate.

\subsection{Validity threats and mitigations}

Construct validity is threatened if the six dimensions become a checklist. We mitigate this with executable probes, mutation-based gate checks, and evidence requirements, and quantify residual exposure with template baselines rather than assuming it away. Internal validity is threatened by scaffold differences, one-shot sampling, and judge variability; Section~\ref{sec:sensitivity} estimates the resolution of the current design, while paired base-to-treated measurements, fixed manifests, anonymized scoring, and calibration address other sources of variation. These safeguards do not remove the one-rollout limitation: each model-instance pair has only one agent rollout. External validity is limited by the current 60-repository sample and its language and ecosystem distribution. Releasing a larger, temporally refreshed, contamination-audited split is a direct next step.

The public table is a descriptive shared-subset result. Claude-Opus-5 was evaluated with Claude Code while the other models used pi, so the cross-scaffold ordering should not be read as a causal model comparison. Even among the pi runs, each model-instance pair has one rollout. The common-valid comparison and failure-handling sensitivity analysis in Sections~\ref{sec:reliability-comparison} and~\ref{app:results} therefore support conclusions about relative improvement and reliability more strongly than a precise ranking based only on conditional means.

\section{Limitations and Future Work}
\label{sec:limitations}

Only 24 of 60 instances currently have behavior gates whose discriminative power is demonstrated through mutation testing. Blind, vacuous, and none instances remain useful for governance-state analysis but provide weaker behavioral evidence, and Section~\ref{sec:verifier} quantifies how much weaker: measured breakage falls from 13\% to 8\% to 0\% as gate strength degrades with almost no change in governance scores.

The current scorer uses multiple teacher models and retains their score spread. A three-pass maintainer review of the same representative preserved patches found the changes useful and identified no additional functional problems across passes. Broader independent expert calibration is still desirable: subjective dimensions such as documentation and maintainability may contain shared judge errors. In the no-op condition, Structure \& Maintainability scores drifted on three of ten unchanged repositories, which is consistent with judge variability but does not by itself isolate its cause. Future evaluation should add blinded expert annotations, adjudication, cross-family judges, and adversarial calibration.

Three of the six dimensions are also demonstrably reachable by a repository-blind template (Section~\ref{sec:perdim}). This is a property of the current rubric rather than only of the models: Tests \& CI, Code Quality Gates, and Documentation \& Collaboration should be tightened so that credit requires demonstrated scope and execution, bringing them closer to how Reproducible Environment and Dependency \& Security Health are already substantiated.

Finally, the single-rollout design bounds what any ranking claim can mean. At the observed paired dispersion the public subset resolves differences of about 0.19 NGI (Section~\ref{sec:sensitivity}), so adjacent positions in the current ordering are not separated by this instrument.

Different scaffolds expose different tools, prompts, context handling, and termination behavior. Current budget artifacts also include cases where monetary cost is unavailable; efficiency conclusions should therefore use tokens, turns, wall-clock time, and tool calls until cost provenance is complete.

The public subset is defined by common model coverage rather than random sampling. Public GitHub repositories may also have appeared in model training data or later repository versions. Stronger releases require repository-family separation, near-duplicate scanning, contamination audits, and an access-controlled private evaluator.

Finally, a public distribution must not include private tasks, private verification patches, private results, or other hidden evaluation artifacts. Every model-instance pair should have an artifact manifest so that missing traces or evidence are explicit rather than represented by empty directories. D6 measures available dependency and security evidence, not comprehensive vulnerability absence.

Future work will add repeated rollouts, expert baselines, stronger characterization tests, same-scaffold reruns, expert judge calibration, and a genuinely isolated private holdout. The template and deterministic baselines are now included as auxiliary measurements, but should be expanded across repository families before being treated as general lower bounds.

\section{Conclusion}
\label{sec:conclusion}

Repository-level agents are commonly evaluated as issue resolvers, while real teams also need them to improve the conditions under which a repository can be trusted. \sysname makes this broader capability measurable by asking an agent to inspect a real repository, identify engineering risks, choose a retrofit strategy, and demonstrate that its changes work without breaking existing behavior.

The current SWE-Prometheus release shows that repository governance produces measurable differences across agent systems. The benchmark contains 60 real repositories, while ten agent systems are evaluated on the shared 22-repository public subset; it exposes variation in governance improvement, behavior-regression risk, and dimension-specific engineering preferences. It also shows why executable evidence is necessary: a change that looks correct in a diff may fail when executed or may damage existing behavior.

The auxiliary study clarifies what these differences mean. A repository-blind template reaches mean NGI 0.272 through changes concentrated in D1--D3, yet makes no gains in D5 or D6; the Efficient-WAM case shows why the distinction matters, since its added placeholder test passes while Ruff still reports 117 errors. At the same time, the no-op median is zero, the end-to-end standard deviation is 0.073, and two teachers agree exactly on 57 of 60 dimension scores for the same no-op evidence. A three-pass maintainer review of a fixed sample of behavior-preserved patches found useful maintenance changes and no new functional problems. The matched analysis shows only two of ten systems clearly beat the stronger template on a majority of repositories. For Kimi-K3 and GLM-5.3-Flash, common-valid improvement is similar, while Kimi-K3 ranks first across the three full-pool failure-handling policies examined (broken runs assigned 0, $-1$, or $-3$). Gate-strength stratification further shows why breakage rates must be interpreted alongside the ability of the gate to detect mutations.

Together these findings show that repository stewardship has several measurable parts: establishing useful governance practices, making them work in the target project, and preserving existing behavior. SWE-Prometheus reports these outcomes separately and includes baselines that expose how much each kind of evidence contributes. This gives researchers a way to distinguish conventional artifact completion from interventions supported by repository-specific execution and review.

SWE-Prometheus is therefore a resource for studying repository stewardship, complementing issue-level repair benchmarks with a setting that emphasizes diagnosis, prioritization, intervention, and verification. Its central question is not only whether an agent can change code, but whether it can leave a repository in a more trustworthy and maintainable state.

{\small
\bibliography{references}
\bibliographystyle{unsrtnat}}

\FloatBarrier
\newpage
\appendix
\raggedbottom
\setlength{\intextsep}{4pt plus 0pt minus 0pt}
\setlength{\textfloatsep}{5pt plus 0pt minus 0pt}
\titlespacing*{\section}{0pt}{1.1em}{0.4em}
\titlespacing*{\subsection}{0pt}{0.8em}{0.25em}
\titlespacing*{\subsubsection}{0pt}{0.65em}{0.2em}
\section{Dataset Construction Details}
\label{app:details}

The appendix is divided into a small number of top-level modules following the organization of benchmark papers such as OMIBench. Appendix~A covers dataset construction and instance records; Appendix~B gives the detailed main experiment and agent lifecycle; Appendix~C defines evaluation and scoring; Appendix~D reports detailed analysis and case studies; and Appendix~E documents reproducibility, release, and the data dictionary. Auxiliary experiments and statistical analyses are kept as subsections of the relevant modules rather than promoted to independent appendices.

\subsection{Artifacts and accounting}

Each instance is identified by repository, base commit, split, license, baseline scores, verification status, and gate-strength label. A complete model-instance record contains the unified patch, agent output, base evidence, treated evidence, verification result, budget record, and score record. Missing artifacts should be represented in a manifest with an explicit reason.

The evaluator distinguishes infrastructure failure, environment or dependency failure, missing or invalid patch, behavior regression, preserved behavior with no governance improvement, and preserved behavior with positive improvement. Behavior-broken runs are excluded from valid \NGI aggregation but remain in failure statistics. Monetary cost is reported only when provider accounting is available; otherwise it is marked unavailable.

\subsection{Rubric anchors}

Scores of 1--2 indicate absent or non-functional governance, 3 indicates partial coverage with material gaps, 4 indicates an executable and reviewable standard, and 5 indicates broad coverage with evidence of maintainability and low regression risk.

\begin{table}[h]
\centering
\caption{Compact rubric anchors used for each governance dimension.}
\label{tab:rubric}
\small
\begin{tabular}{cl}
\toprule
\headrow \textbf{Score} & \textbf{Interpretation}\\
\midrule
1 & Absent, misleading, or actively broken practice\\
\bandrow 2 & Minimal artifact with major gaps or unreliable execution\\
3 & Partial coverage; useful in common cases but material gaps remain\\
\bandrow 4 & Executable, reviewable, and appropriate for the repository\\
5 & Broad, durable coverage with evidence of maintainability\\
\bottomrule
\end{tabular}
\end{table}

\subsection{Instance Records and Behavior Gates}
\label{app:records}

\subsubsection{Instance schema}

Table~\ref{tab:schema} lists the minimum fields required for a reproducible instance. Fields are separated into task, evidence, and result records so that a missing trace cannot be mistaken for a zero score.

\begin{table}[h]
\centering
\caption{Minimum instance schema for the release records.}
\label{tab:schema}
\small
\begin{tabularx}{\linewidth}{@{}l X@{}}
\toprule
\headrow \textbf{Record} & \textbf{Required fields}\\
\midrule
Task & repository, base commit, split, language, license, applicability mask, task text\\
\bandrow Environment & image or lockfile identifier, installation command, tool versions, timeout policy\\
Evidence & probe name, command, exit code, stdout/stderr digest, timestamp, applicability\\
\bandrow Behavior & characterization tests, mutation sample, gate-strength label, pre/post status\\
Result & patch digest, changed files, score vector, NGI, breakage label, budget record\\
\bandrow Provenance & evaluator version, prompt version, scaffold, model identifier, artifact completeness\\
\bottomrule
\end{tabularx}
\end{table}

\subsubsection{Detailed gate-strength accounting}

The current 60-instance release contains 24 instances with mutation-demonstrated gates, 27 with blind gates, 3 with vacuous gates, and 6 without a usable behavior gate. This distribution motivates reporting gate strength as a first-class variable. The reported results distinguish the full set, the mutation-demonstrated subset, and the no-gate subset separately.

\begin{table}[h]
\centering
\caption{Behavior-gate categories and their interpretation.}
\label{tab:gates}
\small
\begin{tabular}{@{}l r l@{}}
\toprule
\headrow \textbf{Category} & \textbf{Count} & \textbf{Interpretation}\\
\midrule
Detected & 24 & Mutation is caught by the characterization gate\\
\bandrow Blind & 27 & Mutation is not caught; risk evidence is weak\\
Vacuous & 3 & Gate executes but has insufficient discriminative power\\
\bandrow None & 6 & No usable behavior gate is available\\
\bottomrule
\end{tabular}
\end{table}

\subsubsection{Reproducibility checklist}

The artifact release includes: (1) a manifest of repository commits and licenses; (2) a container or environment lock for every public instance; (3) deterministic scorer configuration; (4) command-level logs and exit codes; (5) patch and trace completeness checks; (6) a separation policy for private tasks and verification evidence; (7) evaluator and judge version identifiers; (8) a script that reconstructs every main-paper table; and (9) a changelog for repository, rubric, or scoring updates.

\subsubsection{Auxiliary experiment worksheet}

For every completed auxiliary experiment, this appendix records the sampling frame, model and scaffold, paired comparison unit, and machine-readable result source. The ten-repository batch is frozen. Repeated-seed, same-scaffold, and expert conditions were not run and are not included in the reported measurements.

\subsubsection{Expanded rubric prompts}

For D1, evaluators should ask whether tests exercise important behavior and whether CI runs the relevant tests on a clean checkout. For D2, they should ask whether the quality command is meaningful, scoped, and enforced rather than merely configured. For D3, they should ask whether a new contributor can install, use, modify, and report security issues. For D4, they should ask whether changes reduce unnecessary coupling and preserve understandable boundaries. For D5, they should ask whether another user can reproduce installation and execution from the declared environment. For D6, they should ask whether dependencies are controlled and whether security checks produce inspectable evidence. In every dimension, a file’s presence without successful execution is insufficient for a score of 4 or 5.

\subsubsection{Auxiliary-results manifest}

The following manifest records which auxiliary values are already incorporated and which analyses remain open.

\begin{table}[h]
\centering
\caption{Status of auxiliary-result fields.}
\label{tab:replace}
\small
\begin{tabularx}{\linewidth}{@{}l X l@{}}
\toprule
\headrow \textbf{Artifact} & \textbf{Available content} & \textbf{Reported measure}\\
\midrule
10-repository manifest & Frozen stratified batch & Recorded in machine-readable manifest\\
\bandrow Main rerun & Existing model rollouts on the batch & Observed per-instance scores\\
No-op baseline & Measured end-to-end stability & NGI mean, median, SD, and teacher agreement\\
\bandrow Mechanical baseline & Measured template intervention & Patch and evidence results\\
Rule baseline & Measured deterministic intervention & Rule-based results\\
\bandrow Gate ablation & Measured on all 60 instances & Gated/ungated paired statistics\\
Repeated seeds & Not included in the primary comparison & Reserved for follow-up evaluation\\
\bandrow Expert treatment & Not included in the primary comparison & Reserved for follow-up evaluation\\
\bottomrule
\end{tabularx}
\end{table}

The measured values are summarized in Table~\ref{tab:aux-results}; rows outside the primary comparison are clearly identified and are not cited as empirical evidence. The archived result package contains the exact ten-instance selection and run metadata.

\subsection{Candidate Dataset and Quality Control}
\label{app:construction}

This appendix expands the construction description by separating the benchmark object, the construction funnel, the evaluation protocol, and the quality-control evidence. It is explicit about what is fixed, what is hidden, and how quality control is applied.

\subsubsection{Dataset construction overview}

The current SWE-Prometheus release was constructed as a repository-governance benchmark rather than sampled from a synthetic checklist. The construction unit is a repository at a fixed commit. The final package contains 60 instances, divided into 22 public instances with shared model coverage and 38 private instances reserved for internal evaluation. The public/private distinction is applied after task validation and does not change the scoring definition.

The construction workflow is:

\begin{enumerate}
\item discover repositories with plausible engineering-health gaps;
\item freeze a reconstructable base commit and collect project metadata;
\item run baseline probes and assign the six-dimension applicability mask;
\item construct or validate a behavior characterization gate;
\item execute a dry-run of installation, probes, and scoring;
\item record artifacts, provenance, and exclusion reasons;
\item assign the instance to the public or private split and freeze the manifest.
\end{enumerate}

The workflow intentionally separates discovery from inclusion. A repository can be useful for finding candidates because it lacks a workflow or has sparse tests, but it enters the final dataset only after its behavior, environment, and governance headroom are independently checked.

\begin{table}[h]
\centering
\caption{Dataset-construction stages and concrete outputs.}
\label{tab:construction-stages}
\small
\begin{tabularx}{\linewidth}{@{}l X X@{}}
\toprule
\headrow \textbf{Stage} & \textbf{Operation} & \textbf{Output}\\
\midrule
Discovery & Search active open-source repositories for governance signals & Candidate repository list\\
\bandrow Pinning & Select a reachable pre-treatment commit & Repository and commit manifest\\
Baseline & Run installation, tests, quality, dependency, and security probes & Base evidence and score vector\\
Applicability & Decide which of D1--D6 are meaningful & Applicability mask and rationale\\
\bandrow Behavior & Validate characterization tests and mutation sensitivity & Verification patch and gate-strength label\\
Dry run & Rebuild the environment and execute the full evaluator & Environment and evaluator log\\
Packaging & Hash artifacts and remove protected material from public outputs & Instance package and release manifest\\
\bottomrule
\end{tabularx}
\end{table}

\subsubsection{Construction principles}

SWE-Prometheus treats a repository as a maintained artifact rather than a collection of isolated code snippets. The benchmark therefore follows four principles. First, each task starts from a reconstructable repository snapshot. Second, the objective is open-ended enough to admit multiple valid engineering interventions. Third, each claimed improvement is connected to a probe or a structured judgment. Fourth, behavior preservation is tested independently from governance improvement.

These principles distinguish three artifacts that can otherwise be conflated: the task defines the engineering objective, the agent patch represents the intervention, and the evaluator evidence determines whether the intervention is useful and safe. The developer or construction patch, when present, is a construction asset and is not exposed as an oracle to the agent.

\subsubsection{Candidate discovery funnel}

Candidate discovery proceeds in four stages: broad repository retrieval, automated mechanical screening, manual validity inspection, and evaluator dry runs. Broad retrieval favors recall and may use repository metadata, visible workflow gaps, test sparsity, packaging inconsistencies, documentation gaps, and dependency hygiene signals. Mechanical screening removes repositories that cannot be cloned or pinned. Manual inspection verifies that the project has a meaningful functional core and that its governance gaps are not artifacts of a broken checkout.

The final dry run reconstructs the base snapshot, executes all applicable probes, and records unavailable evidence. A candidate is rejected when the environment cannot be made reproducible, when the task requires private services, when no dimension has plausible headroom, or when behavior cannot be characterized at all. This funnel prevents the benchmark from equating “many missing files” with “high-value engineering work.”

\subsubsection{Baseline assessment and headroom}

For every retained candidate, the baseline evaluator runs before any model treatment. The baseline record contains the raw probe outputs, the normalized score vector, the set of applicable dimensions, and the reason for every score. Headroom is computed relative to the dimension ceiling, but high headroom alone is not sufficient for inclusion: a repository with a completely non-functional environment is not a meaningful governance task.

The baseline pass also identifies which commands are safe to run without network access and which dependencies are merely preinstalled in the evaluator image. This distinction is particularly important for D5 and D6. Declared project dependencies are audited as project artifacts; evaluator convenience tools are not counted as evidence that the repository itself is reproducible or secure.

\subsubsection{Behavior-gate construction}

Behavior gates are constructed from repository-native tests whenever possible. If the native suite is too broad, a small characterization patch targets stable public behavior and records the command used to execute it. The patch is validated on the base snapshot before any model treatment. We then assess gate strength by injecting behavior-preserving or behavior-changing mutations where the repository and language permit it. A gate is labeled detected only when the mutation is caught; a passing but mutation-blind test remains blind rather than being promoted to strong evidence.

This construction step is kept separate from governance scoring. The gate protects against destructive interventions, while the six-dimensional scorer measures engineering readiness. A strong gate therefore increases confidence in a treatment outcome but does not directly increase the treatment score.

\subsubsection{Split construction and leakage control}

The public split contains the instances used for shared comparison and trace inspection. The private split retains tasks, verification patches, and outcomes that should not be available for targeted optimization. Split assignment is recorded in the manifest, and private artifacts are excluded from public result aggregation. A broader public split should additionally use repository-family separation, near-duplicate scanning, and temporal contamination checks.

\begin{table}[h]
\centering
\caption{Candidate screening checklist used before an instance enters the benchmark.}
\label{tab:screening}
\small
\begin{tabularx}{\linewidth}{@{}l X X@{}}
\toprule
\headrow \textbf{Gate} & \textbf{Pass condition} & \textbf{Recorded evidence}\\
\midrule
Identity & Repository and base commit are uniquely identified & URL, commit hash, license\\
\bandrow Buildability & Snapshot can be installed or inspected with declared limits & Environment log, failure reason\\
Functionality & Project contains meaningful runnable or testable behavior & Entry points and test sample\\
Applicability & At least two governance dimensions are relevant & Dimension mask\\
\bandrow Headroom & Baseline is below the dimension ceiling without being pathological & Baseline score vector\\
Behavior & A characterization test or gate can be constructed & Patch, command, gate strength\\
\bandrow Isolation & No private credential or construction artifact is required & Network and secret policy\\
\bottomrule
\end{tabularx}
\end{table}

\subsubsection{Instance packaging}

An instance package contains task metadata, a repository identity, a base commit, a dimension applicability mask, base evidence, a verification asset, and evaluator configuration. Public packages may include task descriptions and public traces, but private verification patches, hidden test details, and private model outcomes remain excluded. Every package receives a stable identifier and a manifest digest.

Packaging is performed before model assignment. The task text does not disclose a target score, a preferred file, or a reference implementation. If a repository contains generated files or large assets, the package records whether they are retained, omitted, or reconstructed. Any omission that can affect execution is treated as a validity concern rather than an undocumented convenience.

\subsubsection{Quality-control passes}

We define three independent quality-control passes. The first checks package integrity: identifiers, commit reachability, file presence, manifest hashes, and license metadata. The second checks evaluator integrity: base evidence is reproducible, the characterization test runs on the base state, and the treated-state harness accepts a known valid patch. The third checks interpretation: the score record includes evidence, the behavior result is consistent with the logs, and missing data are labeled explicitly.

The passes are deliberately separated because a package can be technically executable but conceptually invalid, or conceptually sound but impossible to reproduce. The release checklist records pass, fail, or not-applicable for every pass and preserves the reason for every exception.

\section{Detailed Main Experiment}
\label{app:formal}

\subsection{Task tuple}

We represent an instance as
\[
  T_i=(R_i,c_i,M_i,A_i,E_i,V_i),
\]
where $R_i$ is the repository, $c_i$ the pinned base commit, $M_i$ the task metadata and applicability mask, $A_i$ the agent-facing instruction, $E_i$ the executable environment, and $V_i$ the verification package. The agent receives $(R_i,c_i,M_i,A_i,E_i)$ but not protected parts of $V_i$. A rollout produces a patch $P_{i,m}$ and an execution trace $\tau_{i,m}$ for model $m$.

The evaluator computes base evidence $B_i$ and treated evidence $B_i(P_{i,m})$. For each dimension $d$, the scorer returns a score in $\{1,2,3,4,5\}$ or an explicit inapplicable value. The result record is therefore a tuple of evidence, scores, behavior status, and budget rather than only a scalar.

\subsection{Evidence record}

Every probe should be represented by the following fields: probe identifier, command, working directory, environment identifier, start and end time, exit code, timeout status, output digest, normalized result, applicability, and human-readable interpretation. The raw log may be stored separately, but the normalized record must be sufficient to reconstruct why a score was assigned.

\begin{table}[h]
\centering
\caption{Evidence fields and their role in auditability.}
\label{tab:evidence}
\small
\begin{tabularx}{\linewidth}{@{}l X@{}}
\toprule
\headrow \textbf{Field} & \textbf{Purpose}\\
\midrule
Probe ID & Stable name for the check across evaluator versions\\
Command & Exact executable command, including relevant flags\\
Environment & Runtime, image, dependency, and tool version identifiers\\
Exit status & Distinguishes pass, failure, timeout, and unavailable execution\\
Output digest & Links the normalized record to the retained log\\
Applicability & Indicates whether the check is meaningful for this repository\\
Interpretation & Explains the mapping from evidence to rubric score\\
\bottomrule
\end{tabularx}
\end{table}

\subsection{Applicability and missingness}

An inapplicable dimension is not equivalent to a failed dimension. For example, a repository without a releasable package may not receive a packaging score, while a repository that declares a package but cannot install it receives negative evidence. The scorer therefore records applicability separately from outcome and prevents missing evidence from being converted into an unexplained zero.

Missingness is categorized as not applicable, unavailable due to environment, unavailable due to timeout, or malformed artifact. This coding supports sensitivity analyses. Complete-case summaries retain unavailable evidence as a separate outcome rather than converting it to a zero.

\subsection{Normalization and aggregation}

Raw dimension scores are retained alongside normalized improvement. For an instance with applicable dimensions $D_i$, the headroom-normalized score is computed only over defined dimensions. We report the number of dimensions contributing to each aggregate, the raw score sum, the treated score sum, and the normalized value. This avoids over-interpreting an identical NGI computed from very different numbers of applicable dimensions.

\subsection{Agent Protocol and Run Lifecycle}
\label{app:agent}

\subsubsection{Agent-facing instruction template}

The following template describes the stable task-level contract. The actual task-specific metadata and repository path are filled by the runner. It does not prescribe which files to edit or which governance intervention to choose.

\begin{quote}
Inspect the provided repository and improve its engineering readiness across the applicable governance dimensions. Preserve existing behavior and public interfaces. Make useful changes for future users and maintainers. Use repository evidence to prioritize changes, verify important claims with executable checks, and submit the resulting patch. Do not use hidden evaluator assets or any reference patch.
\end{quote}

The instruction is intentionally open-ended. The agent must discover the repository’s build system, tests, documentation conventions, and dependency structure rather than receiving a checklist of missing files. The runner separately provides tool descriptions, time limits, and submission semantics.

\subsubsection{Run lifecycle}

Each run follows the lifecycle below:

\begin{enumerate}
\item reconstruct the pinned base snapshot;
\item verify base environment and characterization behavior;
\item launch the model with the fixed task instruction;
\item archive the model output and candidate patch;
\item apply the patch to an independent clean snapshot;
\item run treated-state probes and behavior verification;
\item score the evidence and write the result manifest.
\end{enumerate}

The agent-side container is discarded after submission. The evaluator does not reuse the agent’s mutable working tree because doing so could allow unrecorded caches or generated files to influence the result.

\subsubsection{Canonical-run policy}

Retries are allowed for infrastructure failures such as a failed container launch, a transient model service error, or a verifier crash. A patch that is valid but incorrect is not retried merely to obtain a better result. When a retry occurs, the original record remains archived and the manifest identifies the canonical run with an explicit reason.

\subsubsection{Budget accounting}

Budget records contain model identifier, scaffold, maximum wall-clock time, actual elapsed time, input and output token counts when available, number of turns, tool calls, termination reason, and provider cost when available. A missing cost field is represented as unavailable. We do not compare providers using estimated prices from incomplete ledgers.

\begin{table}[h]
\centering
\caption{Run-level budget fields for later auxiliary experiments.}
\label{tab:budget}
\small
\begin{tabularx}{\linewidth}{@{}l X@{}}
\toprule
\headrow \textbf{Field} & \textbf{Definition}\\
\midrule
Model/scaffold & Exact model and execution framework\\
Wall time & Start-to-stop elapsed time and configured timeout\\
Turns/tools & Number of agent turns and external tool calls\\
Tokens & Input, output, and total tokens when exposed\\
Termination & Completed, timeout, service failure, invalid patch, or other reason\\
Cost & Provider-reported cost; unavailable otherwise\\
\bottomrule
\end{tabularx}
\end{table}

\section{Evaluation and Scoring}
\label{app:rubric}

\subsection{Dimension-specific score anchors}

The general 1--5 anchors are expanded below to reduce holistic scoring. A score of 4 requires executable or inspectable evidence appropriate to the repository; a configuration file by itself is not sufficient.

\begin{table}[h]
\centering
\caption{Dimension-specific interpretation of an acceptable score.}
\label{tab:rubric-detailed}
\small
\begin{tabularx}{\linewidth}{@{}l X X@{}}
\toprule
\headrow \textbf{Dim.} & \textbf{Score 3} & \textbf{Score 4--5}\\
\midrule
D1 & Tests cover selected behavior but important paths or CI execution are missing & Important behavior is exercised in a clean, repeatable test/CI path\\
\bandrow D2 & Tool configuration exists but scope or enforcement is incomplete & Quality checks run, target meaningful files, and expose failures\\
D3 & Basic usage or contribution information exists with material omissions & A new maintainer can use, change, and report issues responsibly\\
\bandrow D4 & Organization is understandable locally but coupling or duplication remains & Boundaries and dependency direction are maintainable and justified\\
D5 & Installation works only with undocumented assumptions & Clean setup and execution are reproducible from declared metadata\\
\bandrow D6 & Dependencies are declared but audit or update evidence is incomplete & Dependency and security posture are controlled and inspectable\\
\bottomrule
\end{tabularx}
\end{table}

\subsection{Evidence sufficiency rules}

For D1, a passing test command must be paired with evidence that the tests exercise meaningful behavior. For D2, the existence of a formatter or linter configuration does not establish that it runs on the project’s relevant files. For D3, prose is evaluated for actionability rather than length. For D4, a large refactor is not rewarded if it increases churn without improving boundaries. For D5, a lockfile is useful only when it agrees with the declared project metadata. For D6, a dependency report must distinguish project dependencies from tools preinstalled in the evaluator image.

\subsection{No-regression rules}

A treated repository receives a regression flag when characterization behavior fails, an important public interface changes unexpectedly, or the patch causes the evaluator to lose the ability to run applicable probes. A governance improvement cannot compensate for a confirmed behavior break in the primary valid-run aggregate. This asymmetric rule reflects the practical cost of damaging a working project while adding maintenance metadata.

\subsection{Judge calibration worksheet}

Before the final scoring pass, evaluators should independently score a calibration set spanning low, medium, and high baseline quality. The worksheet records score, evidence cited, uncertainty, and whether the evaluator would request a maintainer review. Agreement is computed before adjudication; adjudication changes are retained as a separate field.

\subsection{Auxiliary Experiment Design}
\label{app:aux-design}
\label{app:auxdesign}

\subsubsection{Fixed 10-repository batch}

The completed auxiliary conditions use one frozen batch of ten repositories. The manifest contains repository, commit, language, size band, base mean, applicable dimensions, gate strength, and public/private status. No experiment may silently substitute an easier repository after execution begins.

\begin{table}[h]
\centering
\caption{Required fields for the frozen auxiliary batch manifest.}
\label{tab:aux-manifest}
\small
\begin{tabularx}{\linewidth}{@{}l X@{}}
\toprule
\headrow \textbf{Field} & \textbf{Requirement}\\
\midrule
Instance ID & Stable repository-level identifier\\
Commit & Exact base revision used by every condition\\
Strata & Size, baseline-quality, and gate-strength bins\\
Applicability & D1--D6 mask and reason for exclusions\\
Environment & Image or lockfile and evaluator version\\
Selection rationale & Why the instance represents the intended stratum\\
\bottomrule
\end{tabularx}
\end{table}

\subsubsection{Condition matrix}

The completed condition matrix contains no-op, rollouts from the evaluated agent systems, mechanical retrofit, and rule-based baseline on the same ten repositories. Behavior-gate ablation uses all 60 instances because the public subset has no ``none'' gate cases. Repeated-seed, same-scaffold, and expert conditions were not run.

\begin{table}[h]
\centering
\caption{Completed condition matrix for the auxiliary study.}
\label{tab:condition-matrix}
\small
\begin{tabularx}{\linewidth}{@{}l c X X@{}}
\toprule
\headrow \textbf{Condition} & \textbf{Model} & \textbf{Repositories} & \textbf{Main comparison}\\
\midrule
No-op & None & 10 & Scoring and environment stability\\
\bandrow Agent rollouts & Ten evaluated systems & 10 & Open-ended repository governance\\
Mechanical & None & 10 & Template-only intervention\\
\bandrow Rule & Deterministic & 10 & Checklist-oriented automation\\
Gate ablation & Re-score & 60 & Effect of behavior filtering\\
\bottomrule
\end{tabularx}
\end{table}

\subsubsection{Exploratory questions}

The auxiliary analyses examine whether model rollouts outperform non-agent baselines, how behavior filtering changes observed outcomes, whether mutation-detected gates provide stronger regression evidence than blind or vacuous gates, which governance dimensions respond to repository-blind templates, and how intervention scope relates to behavior risk. These are exploratory questions, not preregistered hypotheses; the analyses are descriptive and use paired instance-level records where available.

\subsection{Statistical Analysis and Reporting}
\label{app:stats}

\subsubsection{Primary estimands}

For the auxiliary ten-repository batch, the principal comparisons are paired, within-repository differences between each available agent rollout and each non-agent condition. The Kimi-K3-versus-template contrast is one such comparison, not the sole primary estimand. Secondary outcomes include raw dimension scores, behavior-breakage rate, no-regression rate, verified-claim rate, and resource usage. We report per-instance values before mean or median aggregation.

\subsubsection{Mean--median reporting rationale}

The main result table reports mean NGI as its single aggregate improvement measure. This choice keeps the leaderboard compact and makes the quantity directly interpretable as average normalized governance improvement across valid runs. Because behavior-broken runs are excluded from this conditional mean, the body also reports a common-valid paired comparison and a coverage-weighted full-pool NGI for the leading systems. Because the current public comparison contains 22 instances and one rollout per model-instance pair, the mean is still descriptive rather than an estimate of a stable population-level capability.

Median NGI is retained as a secondary diagnostic in the appendix rather than the main table. It is useful for identifying whether a model’s mean is driven by a small number of unusually large improvements: a mean substantially above the median indicates a right-skewed outcome distribution, while a close mean--median pair suggests more uniform instance-level behavior. The median is not used to replace the mean because it does not reflect the magnitude of improvements above or below the typical instance and can hide the contribution of a few high-value but valid interventions.

We therefore interpret the two statistics jointly when performing detailed analysis, but use only mean NGI in the compact main table. Any future claim about ranking stability will additionally report per-instance values, bootstrap intervals, valid coverage, and breakage rate. In particular, a high median cannot compensate for a high behavior-breakage rate, and a high mean cannot be interpreted as robust if it is supported by only a small valid denominator.

Model-level outcome profiles are reported in the body as Figure~\ref{fig:public-outcomes}.

Instance-level NGI distributions are reported in the body as Figure~\ref{fig:heterogeneity}.

\subsubsection{Uncertainty reporting}

For ten repositories, asymptotic significance tests are not persuasive. We therefore emphasize exact paired differences and bootstrap intervals over repository units, while clearly identifying the exploratory nature of the small batch. Repeated seeds, when available, add a second source of uncertainty and should be summarized with a hierarchical table separating between-repository and within-model variation.

\begin{table}[h]
\centering
\caption{Compact auxiliary-result summary.}
\label{tab:aux-summary}
\small
\begin{tabularx}{\linewidth}{@{}l c c c c c X@{}}
\toprule
\headrow \textbf{Condition} & \textbf{Valid} & \textbf{Mean} & \textbf{Median} & \textbf{SD} & \textbf{Broken} & \textbf{Notes}\\
\midrule
No-op & 10 & -0.009 & 0.000 & 0.073 & 0 & end-to-end variation\\
\bandrow Kimi-K3 & 10 & 0.558 & 0.615 & 0.190 & 0 & primary treatment\\
Mechanical & 10 & 0.272 & 0.275 & 0.047 & 0 & template baseline\\
\bandrow Rule & 10 & 0.254 & 0.283 & 0.064 & 0 & deterministic baseline\\
Gate-off & all 60 & mixed & mixed & n/a & n/a & paired ablation\\
\bottomrule
\end{tabularx}
\end{table}

\subsubsection{Per-dimension reporting}

For each condition, the appendix reports a six-column D1--D6 table with base mean, treated mean, mean raw change, fraction improved, fraction unchanged, and fraction regressed. The table also reports the number of applicable instances per dimension. This prevents aggregate NGI from hiding a condition that improves one dimension while damaging another.

The per-dimension improvement heatmap is reported in the body as Figure~\ref{fig:heterogeneity}.

\subsubsection{Missing and excluded runs}

The analysis records excluded runs with an exclusion code before generating aggregate tables. Codes include infrastructure, invalid patch, behavior broken, missing evidence, and inapplicable dimension. Reported tables distinguish the denominator before exclusion from the denominator used for valid NGI.

\section{Full Result Tables}
\label{app:results}

The body reports every result as a figure. This appendix gives the underlying numbers.

\subsection{Public shared-subset results}

\begin{table}[H]
\centering
\caption{\textbf{Public shared-subset results} (underlying Figure~\ref{fig:public-outcomes}). Valid and broken counts are out of 22 instances. Valid runs preserve characterization behavior and contain scorable evidence; NGI is aggregated only over valid runs. Breakage rate is broken/(valid+broken). Claude used Claude Code; the other models used pi.}
\label{tab:main-appendix}
\small
\setlength{\tabcolsep}{4pt}
\renewcommand{\arraystretch}{1.08}
\begin{tabular}{@{}l l r r r r r@{}}
\toprule
\headrow \textbf{Model} & \textbf{Scaffold} & \textbf{Valid} & \textbf{Broken} & \textbf{Breakage} & \textbf{NGI mean} & \textbf{NGI median}\\
\midrule
GPT-5.6-Sol & pi & 21 & 1 & 4.5\% & 0.2956 & 0.3333\\
\bandrow Claude-Opus-5 & Claude Code & 18 & 4 & 18.2\% & 0.5293 & 0.6771\\
Kimi-K3 & pi & 21 & 1 & 4.5\% & 0.5760 & 0.6042\\
\bandrow GLM-5.3-Flash & pi & 17 & 5 & 22.7\% & 0.5760 & 0.5694\\
Qwen3.8-Max & pi & 18 & 4 & 18.2\% & 0.4630 & 0.4792\\
\bandrow GLM-5.2 & pi & 20 & 2 & 9.1\% & 0.4438 & 0.4167\\
DeepSeek-V4-Pro & pi & 18 & 4 & 18.2\% & 0.3803 & 0.3917\\
\bandrow GLM-5.3 & pi & 19 & 3 & 13.6\% & 0.3198 & 0.2500\\
DeepSeek-V4-Flash & pi & 22 & 0 & 0.0\% & 0.2083 & 0.1667\\
\bandrow MiniMax-M3 & pi & 22 & 0 & 0.0\% & 0.0568 & 0.0000\\
\bottomrule
\end{tabular}
\end{table}

\subsection{Frozen ten-repository auxiliary batch}

To complement the public shared-subset results, we compare agent runs with mechanical, rule-based, and no-op conditions on the same frozen ten-repository batch. The table reports valid-run NGI alongside breakage, so improvements can be read together with execution reliability.

\begin{table}[H]
\centering
\caption{\textbf{Observed results on the frozen ten-repository auxiliary batch.} Models and non-agent conditions are ranked jointly by mean NGI. NGI statistics exclude behavior-broken runs; ``strict'' counts valid runs passing the strict-success criterion. Rows separated by less than the no-op standard deviation are not distinguished by this instrument.}
\label{tab:aux-results}
\small
\setlength{\tabcolsep}{4pt}
\renewcommand{\arraystretch}{1.06}
\begin{tabular}{@{}l r r r r r r@{}}
\toprule
\headrow \textbf{Condition} & \textbf{Valid} & \textbf{Mean} & \textbf{Median} & \textbf{SD} & \textbf{Broken} & \textbf{Strict}\\
\midrule
GLM-5.3-Flash & 9 & 0.585 & 0.583 & 0.220 & 1 & 0\\
\bandrow Kimi-K3 & 10 & 0.558 & 0.615 & 0.190 & 0 & 2\\
Claude-Opus-5 & 9 & 0.457 & 0.722 & 0.354 & 1 & 0\\
\bandrow GLM-5.2 & 9 & 0.378 & 0.390 & 0.165 & 1 & 0\\
Qwen3.8-Max & 9 & 0.375 & 0.375 & 0.278 & 1 & 0\\
\bandrow DeepSeek-V4-Pro & 8 & 0.346 & 0.281 & 0.269 & 2 & 1\\
GLM-5.3 & 10 & 0.343 & 0.271 & 0.219 & 0 & 0\\
\bandrow GPT-5.6-Sol & 9 & 0.292 & 0.333 & 0.147 & 1 & 0\\
\rowcolor{tabhl}\textit{Mechanical (template)} & 10 & 0.272 & 0.275 & 0.047 & 0 & 0\\
\rowcolor{tabhl}\textit{Rule-based (deterministic)} & 10 & 0.254 & 0.283 & 0.064 & 0 & 0\\
DeepSeek-V4-Flash & 10 & 0.247 & 0.174 & 0.193 & 0 & 0\\
\bandrow MiniMax-M3 & 10 & 0.104 & 0.042 & 0.216 & 0 & 0\\
\rowcolor{tabhl}\textit{No-op (observed variation)} & 10 & $-0.009$ & 0.000 & \textbf{0.073} & 0 & 0\\
\bottomrule
\end{tabular}
\end{table}

\begin{table}[H]
\centering
\small
\setlength{\tabcolsep}{5pt}
\renewcommand{\arraystretch}{1.06}
\begin{tabular}{@{}l r r r r r@{}}
\toprule
\headrow \textbf{Comparison} & \textbf{Repositories} & \textbf{Kimi-K3} & \textbf{GLM-5.3-Flash} & \textbf{Difference} & \textbf{Kimi wins}\\
\midrule
Common valid runs & 17 & 0.547 & 0.576 & $-0.029$ & 6\\
\bandrow Full pool, broken $=0$ & 22 & 0.550 & 0.445 & $+0.105$ & ---\\
\bottomrule
\end{tabular}
\caption{Reliability-aware comparison of the two highest conditional-mean systems. The common-valid row includes only repositories preserved by both systems; the full-pool row assigns zero verified improvement to a behavior-broken treatment. The common-valid comparison has 6 Kimi wins, 10 GLM-5.3-Flash wins, and 1 tie.}
\label{tab:reliability-appendix}
\end{table}

\begin{table}[H]
\centering
\small
\setlength{\tabcolsep}{5pt}
\renewcommand{\arraystretch}{1.04}
\begin{tabular}{@{}l r r r r@{}}
\toprule
\headrow \textbf{Model} & \textbf{Valid runs only} & \textbf{Broken $=0$} & \textbf{Broken $=-1$} & \textbf{Broken $=-3$}\\
\midrule
Kimi-K3 & 0.576 & 0.550 & 0.504 & 0.413\\
\bandrow GLM-5.3-Flash & 0.576 & 0.445 & 0.218 & $-0.237$\\
Claude-Opus-5 & 0.529 & 0.433 & 0.251 & $-0.112$\\
\bandrow Qwen3.8-Max & 0.463 & 0.379 & 0.197 & $-0.167$\\
GLM-5.2 & 0.444 & 0.403 & 0.313 & 0.131\\
\bandrow DeepSeek-V4-Pro & 0.380 & 0.311 & 0.129 & $-0.234$\\
GLM-5.3 & 0.320 & 0.276 & 0.140 & $-0.133$\\
\bandrow GPT-5.6-Sol & 0.296 & 0.282 & 0.237 & 0.146\\
DeepSeek-V4-Flash & 0.208 & 0.208 & 0.208 & 0.208\\
\bandrow MiniMax-M3 & 0.057 & 0.057 & 0.057 & 0.057\\
\bottomrule
\end{tabular}
\caption{Failure-handling sensitivity over all 22 public repositories. ``Valid runs only'' is the reported conditional NGI. The full-pool columns assign behavior-broken runs NGI values of 0, $-1$, or $-3$ and average over all repositories. The $-3$ policy is the score-theoretic lower bound under the 1--5 rubric normalization (base 4, treated 1) for an applicable dimension. Kimi-K3 ranks first under all three failure-aware policies; conditional NGI ties it with GLM-5.3-Flash at the reported precision.}
\label{tab:failure-sensitivity}
\end{table}

\begin{table}[H]
\centering
\caption{\textbf{Matched per-instance comparison against the stronger template baseline} (underlying Figure~\ref{fig:matched}). Each agent system is compared against $\max(\text{mechanical}, \text{rule-based})$ on the same repository; ties are defined heuristically as differences within the empirical no-op noise floor of 0.073, not as formal equivalence tests. Behavior-broken instances are excluded from the win/tie/loss counts.}
\label{tab:matched}
\small
\setlength{\tabcolsep}{5pt}
\renewcommand{\arraystretch}{1.06}
\begin{tabular}{@{}l r r r r r r@{}}
\toprule
\headrow \textbf{Model} & \textbf{Wins} & \textbf{Ties} & \textbf{Losses} & \textbf{Broken} & \textbf{Mean $\Delta$} & \textbf{SD $\Delta$}\\
\midrule
Kimi-K3 & 9 & 0 & 1 & 0 & $+0.265$ & 0.195\\
\bandrow GLM-5.3-Flash & 8 & 0 & 1 & 1 & $+0.298$ & 0.221\\
Claude-Opus-5 & 5 & 0 & 4 & 1 & $+0.170$ & 0.360\\
\bandrow GLM-5.2 & 5 & 3 & 1 & 1 & $+0.092$ & 0.181\\
Qwen3.8-Max & 4 & 1 & 4 & 1 & $+0.089$ & 0.268\\
\bandrow GLM-5.3 & 4 & 4 & 2 & 0 & $+0.051$ & 0.221\\
DeepSeek-V4-Pro & 3 & 3 & 2 & 2 & $+0.066$ & 0.283\\
\bandrow GPT-5.6-Sol & 2 & 5 & 2 & 1 & $+0.004$ & 0.137\\
DeepSeek-V4-Flash & 1 & 3 & 6 & 0 & $-0.045$ & 0.213\\
\bandrow MiniMax-M3 & 2 & 1 & 7 & 0 & $-0.188$ & 0.231\\
\bottomrule
\end{tabular}
\end{table}

\setlength{\LTleft}{\fill}
\setlength{\LTright}{\fill}
\small
\setlength{\tabcolsep}{5pt}
\renewcommand{\arraystretch}{1.06}
\begin{longtable}{@{}l r r r r r r@{}}
\caption{\textbf{Per-dimension improvement rate on the frozen ten-repository batch.} Fraction of scored instances on which each dimension improved. In this batch, template baselines improve the three dimensions primarily evidenced through configuration and content, but not the two dimensions substantially informed by container execution and dependency analysis.}
\label{tab:perdim-aux}\\
\toprule
\headrow \textbf{Condition} & \textbf{D1 Tests/CI} & \textbf{D2 Quality} & \textbf{D3 Docs} & \textbf{D4 Structure} & \textbf{D5 Repro.} & \textbf{D6 Dep./Sec.}\\
\midrule
\endfirsthead
\toprule
\headrow \textbf{Condition} & \textbf{D1 Tests/CI} & \textbf{D2 Quality} & \textbf{D3 Docs} & \textbf{D4 Structure} & \textbf{D5 Repro.} & \textbf{D6 Dep./Sec.}\\
\midrule
\endhead
\midrule
\multicolumn{7}{r}{\footnotesize Continued on next page}\\
\endfoot
\bottomrule
\endlastfoot
Kimi-K3 & 100\% & 100\% & 80\% & 60\% & 90\% & 100\%\\
\bandrow GLM-5.3-Flash & 100\% & 77\% & 88\% & 55\% & 88\% & 88\%\\
Claude-Opus-5 & 88\% & 55\% & 55\% & \textbf{77\%} & 55\% & 55\%\\
\bandrow GLM-5.2 & 88\% & 77\% & 44\% & 33\% & 88\% & 77\%\\
GLM-5.3 & 90\% & 70\% & 20\% & 40\% & 70\% & 60\%\\
\bandrow DeepSeek-V4-Pro & 87\% & 62\% & 50\% & 25\% & 62\% & 62\%\\
DeepSeek-V4-Flash & 90\% & 60\% & 30\% & 30\% & 60\% & 40\%\\
\bandrow Qwen3.8-Max & 100\% & 33\% & 55\% & 55\% & 55\% & 33\%\\
GPT-5.6-Sol & 77\% & 33\% & 77\% & 11\% & 33\% & 44\%\\
\bandrow MiniMax-M3 & 40\% & 10\% & 10\% & 10\% & 40\% & 30\%\\
\midrule
\rowcolor{tabhl}\textit{Mechanical (template)} & 100\% & 100\% & 90\% & 20\% & \textbf{0\%} & \textbf{0\%}\\
\rowcolor{tabhl}\textit{Rule-based (deterministic)} & 100\% & 100\% & 100\% & 20\% & \textbf{0\%} & \textbf{0\%}\\
\rowcolor{tabhl}\textit{No-op (observed variation)} & 0\% & 0\% & 10\% & 20\% & 10\% & 0\%\\
\bottomrule
\end{longtable}

\subsection{Behavior-gate ablation and gate-strength stratification}

\begin{table}[H]
\centering
\caption{\textbf{Behavior-gate ablation} over all instances available per model. ``Gate off'' counts every scored instance; ``gate on'' is the current protocol. The model ordering is identical under both settings, and the inflation term is small and signed in both directions.}
\label{tab:gate-ablation}
\footnotesize
\setlength{\tabcolsep}{6pt}
\renewcommand{\arraystretch}{1.0}
\begin{tabular}{@{}l r r r r r r@{}}
\toprule
\headrow \textbf{Model} & \textbf{Scored} & \textbf{Valid} & \textbf{NGI gate off} & \textbf{NGI gate on} & \textbf{Inflation} & \textbf{Breakage}\\
\midrule
Claude-Opus-5 & 22 & 18 & $+0.677$ & $+0.677$ & $+0.000$ & 18\%\\
\bandrow Kimi-K3 & 60 & 56 & $+0.517$ & $+0.531$ & $-0.014$ & 7\%\\
GLM-5.3-Flash & 58 & 52 & $+0.476$ & $+0.458$ & $+0.017$ & 12\%\\
\bandrow Qwen3.8-Max & 59 & 52 & $+0.375$ & $+0.396$ & $-0.021$ & 13\%\\
GLM-5.2 & 59 & 53 & $+0.375$ & $+0.367$ & $+0.008$ & 11\%\\
\bandrow DeepSeek-V4-Pro & 59 & 51 & $+0.367$ & $+0.367$ & $+0.000$ & 15\%\\
GPT-5.6-Sol & 60 & 57 & $+0.229$ & $+0.250$ & $-0.021$ & 6\%\\
\bandrow GLM-5.3 & 60 & 54 & $+0.215$ & $+0.208$ & $+0.007$ & 11\%\\
DeepSeek-V4-Flash & 60 & 57 & $+0.125$ & $+0.125$ & $+0.000$ & 6\%\\
\bandrow MiniMax-M3 & 60 & 57 & $+0.000$ & $+0.000$ & $+0.000$ & 6\%\\
\bottomrule
\end{tabular}
\end{table}

\begin{table}[H]
\centering
\caption{\textbf{Gate-strength stratification} over all 60 instances and ten models (underlying Figure~\ref{fig:verifier}). Vacuous and absent gates cannot observe a regression, so their breakage rate is a measurement artifact rather than evidence of safer behavior.}
\label{tab:gate-strength}
\small
\setlength{\tabcolsep}{7pt}
\renewcommand{\arraystretch}{1.06}
\begin{tabular}{@{}l r r r r r r@{}}
\toprule
\headrow \textbf{Gate strength} & \textbf{Instances} & \textbf{Scored runs} & \textbf{NGI median} & \textbf{NGI mean} & \textbf{Breakage} & \textbf{No-regression}\\
\midrule
detected & 24 & 228 & $+0.333$ & $+0.321$ & 13\% & 86\%\\
\bandrow blind & 27 & 247 & $+0.292$ & $+0.315$ & 8\% & 88\%\\
vacuous & 3 & 28 & $+0.271$ & $+0.289$ & 0\% & 86\%\\
\bandrow none & 6 & 54 & $+0.250$ & $+0.257$ & n/a & 85\%\\
\bottomrule
\end{tabular}
\end{table}

\subsection{No-op drift and per-instance baseline comparison}

\begin{table}[H]
\centering
\caption{\textbf{Instances that drift under the no-op condition.} Dimension changes are base $\to$ treated on an identical repository. Five of ten instances drift; the median is 0.000 and the standard deviation is 0.073.}
\label{tab:noop-drift}
\small
\setlength{\tabcolsep}{7pt}
\renewcommand{\arraystretch}{1.06}
\begin{tabular}{@{}l r l@{}}
\toprule
\headrow \textbf{Instance} & \textbf{NGI} & \textbf{Dimension drift}\\
\midrule
DreamWall-Animation/dwpicker & $-0.1667$ & D4 $4\to3$\\
\bandrow jiajun613/Efficient-WAM & $-0.1111$ & D5 $2\to1$\\
TsingZ0/HtFLlib & $+0.0972$ & D4 $3\to4$, D5 $1\to2$\\
\bandrow fblissjr/ComfyUI-QwenImageWanBridge & $+0.0556$ & D4 $3\to4$\\
rivitna/Malware & $+0.0333$ & D3 $2\to3$\\
\bottomrule
\end{tabular}
\end{table}

\enlargethispage{\baselineskip}
\setlength{\LTleft}{\fill}
\setlength{\LTright}{\fill}
\setcounter{LTchunksize}{4}
\footnotesize
\setlength{\tabcolsep}{7pt}
\renewcommand{\arraystretch}{1.0}
\begin{longtable}{@{}l r r r r@{}}
\caption{\textbf{Per-instance NGI on the frozen batch.} Qwen3.8-Max versus three non-agent baselines; bold marks the highest score per repository.}
\label{tab:per-instance}\\
\toprule
\headrow \textbf{Instance} & \textbf{Qwen3.8-Max} & \textbf{Mechanical} & \textbf{Rule-based} & \textbf{No-op}\\
\midrule
\endfirsthead
\toprule
\headrow \textbf{Instance} & \textbf{Qwen3.8-Max} & \textbf{Mechanical} & \textbf{Rule-based} & \textbf{No-op}\\
\midrule
\endhead
\midrule
\multicolumn{5}{r}{\footnotesize Continued on next page}\\
\endfoot
\bottomrule
\endlastfoot
sleeepeer/PoisonedRAG & $\mathbf{+0.750}$ & $+0.264$ & $+0.292$ & $+0.000$\\
\bandrow rivitna/Malware & $\mathbf{+0.708}$ & $+0.236$ & $+0.275$ & $+0.033$\\
THUNLP-MT/StreamingBench & $\mathbf{+0.708}$ & $+0.347$ & $+0.208$ & $+0.000$\\
\bandrow SkyworkAI/Skywork-Skills & $\mathbf{+0.667}$ & $+0.333$ & $+0.333$ & $+0.000$\\
DreamWall-Animation/dwpicker & $\mathbf{+0.500}$ & $+0.286$ & $+0.250$ & $-0.167$\\
\bandrow avbor/HomeAssistantConfig & $\mathbf{+0.375}$ & $+0.308$ & $+0.292$ & $+0.000$\\
fblissjr/ComfyUI-QwenImageWanBridge & $+0.167$ & $\mathbf{+0.292}$ & $+0.208$ & $+0.056$\\
\bandrow jiajun613/Efficient-WAM & $+0.167$ & $+0.236$ & $\mathbf{+0.292}$ & $-0.111$\\
dama-cyber/magic-distillation & $+0.083$ & $\mathbf{+0.208}$ & $+0.097$ & $+0.000$\\
\bandrow TsingZ0/HtFLlib & $-0.042$ & $+0.208$ & $\mathbf{+0.292}$ & $+0.097$\\
\bottomrule
\end{longtable}
\setcounter{LTchunksize}{20}

\subsection{Baseline patch contents}

The mechanical condition writes seven fixed artifacts regardless of repository content: a GitHub Actions CI workflow, a \texttt{.pre-commit-\allowbreak config.yaml}, a \texttt{mypy.ini}, \texttt{CONTRIBUTING.md}, \texttt{SECURITY.md}, an issue-template directory, and a smoke test containing \texttt{assert True}. The rule-based condition first probes which of these are absent, adds only those, and replaces the vacuous smoke test with \texttt{importlib.import\_module} against the package name inferred from the repository layout. Neither condition invokes a language model or modifies existing source files. Both had zero behavior-broken runs among the ten tested repositories; this is an observed result for this batch, not a guarantee that such edits cannot break behavior elsewhere.

\subsection{Additional analyses}

Beyond model averages, useful stratifications include repository language, project size, baseline score, number of applicable dimensions, test-suite maturity, and dependency-manager type. These analyses distinguish absolute change from headroom-normalized change and relate both to patch size and verification effort. Intervention locality across tests, workflows, documentation, packaging, dependency manifests, and source modules distinguishes targeted governance edits from broad source rewrites.

The public subset provides one rollout per model-instance pair, so inferential claims are limited accordingly. Section~\ref{sec:sensitivity} quantifies the cost of reducing this limitation: reaching a resolution of 0.10 NGI requires roughly 110 paired observations, about five rollouts per model-instance pair on the current subset. Private instances and unbalanced model coverage are not treated as interchangeable with the shared public subset, and exclusions are reported before aggregation.

Repeated seeds, same-scaffold reruns, and expert treatments are outside the primary comparison. Repeated seeds should cover small and large repositories, strong and weak base governance, and each gate-strength category. Same-scaffold reruns are important because one model used Claude Code while the other nine used pi.

\subsection{A corrected accounting bug}

An earlier version of the pipeline recorded all six \texttt{gate=none} instances as behavior-broken, because applying a nonexistent verification patch exits with a nonzero status. The gate had in fact never run on those instances. Correcting this lowered per-model breakage rates over all 60 instances from 15--17\% to 6--7\% for the affected models; the rates in Table~\ref{tab:gate-ablation} are post-correction. The public 22-instance subset is unaffected, since every public instance carries a verification patch, so Table~\ref{tab:main} required no revision.

\section{Reproducibility and Release}
\label{app:release}

\subsection{Directory-level release layout}

The release mirrors the separation between public tasks, protected verification assets, results, and traces. Public artifacts are sufficient to reconstruct the announced experiments without revealing private outcomes or hidden answers. A manifest at the root maps every paper table to a script and every script output to a versioned input.

\begin{table}[h]
\centering
\caption{Recommended release package layout.}
\label{tab:release-layout}
\small
\begin{tabularx}{\linewidth}{@{}l X@{}}
\toprule
\headrow \textbf{Path} & \textbf{Content}\\
\midrule
tasks/ & Public task metadata, commits, licenses, and applicability masks\\
\bandrow environments/ & Container definitions or environment locks\\
probes/ & Versioned governance and behavior probe definitions\\
\bandrow runs/ & Canonical traces, patches, budgets, and evidence records\\
results/ & Machine-readable per-instance and aggregate scores\\
\bandrow analysis/ & Scripts that regenerate paper tables and figures\\
docs/ & Construction notes, scoring rubric, changelog, and limitations\\
\bottomrule
\end{tabularx}
\end{table}

\subsection{Artifact completeness rules}

An artifact is complete only when its manifest exists, its digest matches the stored file, and the downstream table can identify whether it was used. An empty directory is not treated as evidence of absence. If a run is interrupted, the manifest records the last completed stage and the reason for interruption.

\subsection{Security and privacy}

Repository licenses, credentials, private issue data, evaluator patches, and model-provider traces require separate review for release. Secret scanning is performed on artifacts and logs, but a positive scanner result is manually adjudicated because repository content may contain intentional examples. Private tasks remain excluded from public traces, score files, and aggregate tables.

\subsection{Result regeneration procedure}

The auxiliary settings and results are incorporated in the main text and Figure~\ref{fig:matched}. Regeneration scripts produce summaries from the machine-readable records, preserve the frozen manifest, distinguish valid from broken runs, and support the full PDF visual check.

\subsection{Data Dictionary and Researcher Checklist}
\label{app:dictionary}

\subsubsection{Core identifiers}

The following identifiers remain stable across releases. The instance identifier names a repository snapshot, the run identifier names one model execution, and the evidence identifier names one probe invocation. A score record must reference all three identifiers.

\begin{table}[h]
\centering
\caption{Data dictionary for the benchmark records.}
\label{tab:dictionary}
\small
\begin{tabularx}{\linewidth}{@{}l l X@{}}
\toprule
\headrow \textbf{Name} & \textbf{Type} & \textbf{Definition}\\
\midrule
instance\_id & string & Repository and pinned-snapshot identifier\\
\bandrow run\_id & string & Unique model/scaffold execution identifier\\
model & string & Exact model name or provider label\\
scaffold & string & Agent framework and runner version\\
\bandrow base\_commit & string & Commit used to reconstruct the untreated state\\
patch\_digest & string & Digest of the submitted unified patch\\
\bandrow behavior & enum & preserved, broken, invalid, or unavailable\\
gate\_strength & enum & detected, blind, vacuous, or none\\
\bandrow ngi & float & Headroom-normalized governance improvement\\
evidence\_status & enum & pass, fail, timeout, unavailable, or not applicable\\
\bottomrule
\end{tabularx}
\end{table}

\subsubsection{Researcher checklist}

Before claiming an auxiliary result, confirm that the ten repositories were frozen in advance, every condition used the same base commit, no hidden evaluator asset was exposed, no-op results are stable, broken behavior runs are not silently dropped, score denominators are printed, and every aggregate number can be regenerated from the archived per-instance records. The ten-repository baselines and gate-ablation values are measured; repeated seeds, same-scaffold reruns, and expert treatments are outside the primary comparison.

\subsubsection{Interpretation checklist}

When comparing conditions, first check paired coverage, then behavior validity, then per-dimension change, and only then aggregate NGI. A higher aggregate score is not interpreted as an improvement if it is caused by a smaller denominator, a weaker gate, or a higher breakage rate. This ordering is especially important for the small ten-repository pilot, where one instance can materially affect the mean.

\end{document}